\documentclass{article}

\usepackage{graphicx}
\graphicspath{{figures/}}
\usepackage{subfigure}

\usepackage{algorithm,algorithmic}

\usepackage[algo2e,linesnumbered]{algorithm2e}
\usepackage{enumitem}
\setlist{nosep,leftmargin=1em}
\usepackage{booktabs}
\usepackage{multirow}

\usepackage{amsmath,amsthm,amssymb}

\usepackage{hyperref}

\usepackage[accepted]{icml2020}

\newcommand{\const}{\mathrm{Const.}}
\newcommand{\T}{{\hspace{-0.25ex}\top\hspace{-0.25ex}}}
\newcommand{\pr}{\mathrm{Pr}}
\newcommand{\bp}{\boldsymbol{p}}
\newcommand{\by}{\boldsymbol{y}}
\newcommand{\beps}{\boldsymbol{\epsilon}}
\newcommand{\bzero}{\boldsymbol{0}}
\newcommand{\bone}{\boldsymbol{1}}
\newcommand{\bbE}{\mathbb{E}}
\newcommand{\bbI}{\mathbb{I}}
\newcommand{\bbR}{\mathbb{R}}
\newcommand{\cN}{\mathcal{N}}
\newcommand{\cS}{\mathcal{S}}
\newcommand{\cX}{\mathcal{X}}
\newcommand{\cY}{\mathcal{Y}}
\newcommand{\fB}{\mathfrak{B}}
\newcommand{\fC}{\mathfrak{C}}
\newcommand{\fO}{\mathfrak{O}}
\newcommand{\Rhat}{\widehat{R}}
\newcommand{\ytil}{\tilde{y}}
\newcommand{\bytil}{\boldsymbol{\tilde{y}}}

\newcommand{\nmb}{{n_\mathrm{b}}}
\newcommand{\Smb}{{\cS_\mathrm{b}}}
\newcommand{\lmb}{{\ell_\mathrm{b}}}
\newcommand{\blmb}{{\boldsymbol{\ell}_\mathrm{b}}}
\newcommand{\Cg}{{\fC_\textrm{good}}}
\newcommand{\Cb}{{\fC_\textrm{bad}}}
\newcommand{\ellb}{{\ell^\leftarrow}}
\newcommand{\Rhatb}{{\Rhat^\leftarrow}}
\newcommand{\bell}{{\boldsymbol{\ell}_{y\mid f(x)}}}
\newcommand{\bellb}{{\boldsymbol{\ell}_{\ytil\mid f(x)}^\leftarrow}}
\newcommand{\bpy}{{\bp_{y\mid x}}}
\newcommand{\bpyt}{{\bp_{\ytil\mid x}}}
\newcommand{\SL}{{_\textrm{SL}}}
\newcommand{\BC}{{_\textrm{BC}}}

\begin{document}

\twocolumn[
\icmltitle{SIGUA: Forgetting May Make Learning with Noisy Labels More Robust}


\begin{icmlauthorlist}
\icmlauthor{Bo Han}{HKBU,RIKEN}
\icmlauthor{Gang Niu}{RIKEN}
\icmlauthor{Xingrui Yu}{UTS}
\icmlauthor{Quanming Yao}{4P}
\icmlauthor{Miao Xu}{RIKEN,UQ}
\icmlauthor{Ivor W.~Tsang}{UTS}
\icmlauthor{Masashi Sugiyama}{RIKEN,UTokyo}
\end{icmlauthorlist}

\icmlaffiliation{HKBU}{Hong Kong Baptist University}
\icmlaffiliation{RIKEN}{RIKEN}
\icmlaffiliation{UTS}{AAII, University of Technology Sydney}
\icmlaffiliation{4P}{4Paradigm Inc. (Hong Kong)}
\icmlaffiliation{UQ}{University of Queensland}
\icmlaffiliation{UTokyo}{The University of Tokyo}

\icmlcorrespondingauthor{Bo Han}{bhanml@comp.hkbu.edu.hk}


\vskip 0.3in
]

\printAffiliationsAndNotice{}

\begin{abstract}
Given data with noisy labels, over-parameterized deep networks can gradually memorize the data, and fit everything in the end.
Although equipped with corrections for noisy labels, many learning methods in this area still suffer overfitting due to \emph{undesired memorization}.
In this paper, to relieve this issue, we propose \emph{stochastic integrated gradient underweighted ascent}~(SIGUA):
in a mini-batch, we adopt gradient descent on good data as usual, and \emph{learning-rate-reduced gradient ascent} on bad data;
the proposal is a \emph{versatile} approach where data goodness or badness is w.r.t.\ desired or undesired memorization given a \emph{base} learning method.
Technically, SIGUA \emph{pulls optimization back for generalization} when their goals conflict with each other; philosophically, SIGUA shows \emph{forgetting undesired memorization can reinforce desired memorization}.
Experiments demonstrate that SIGUA successfully robustifies two typical base learning methods, so that their performance is often significantly improved.
\end{abstract}

\section{Introduction}
\label{sec:intro}

Data labeling may be heavily noisy in practice where \emph{label generation/corruption processes} are usually agnostic \citep{xiao2015learning,jiang2017mentornet,wang2019symmetric,welinder2010multidimensional}.
As a result, learning with noisy labels seems inevitable.
On the other hand, more complex data requires more expressive power, and then using \emph{over-parameterized deep networks} as our models seems also inevitable \citep{goodfellow2016deep}.
This combination of noisy labels and deep networks is very pessimistic, since deep networks are able to fit anything given for training even if the labels are completely random \citep{zhang2016understanding}.
Unfortunately, it is non-trivial to apply \emph{general-purpose regularization} such as \emph{weight decay} \citep{krogh1991weightdecay} and  \emph{dropout} \citep{srivastava2014dropout} for controlling model complexities of deep networks.
General-purpose regularization would hurt the capability of memorizing not only noisy labels but also complex data, which is never our desideratum.

Fortunately, even though deep networks can fit everything in the end, they \emph{learn patterns first} \citep{arpit2017closer}:
this suggests deep networks can gradually memorize the data, moving from regular data to irregular data such as outliers and mislabeled data.
As a consequence, the memorization events during training may be divided into two categories:
\emph{desired memorization} that helps generalization, and \emph{undesired memorization} that hurts generalization.
Our desideratum is to keep the former and avoid the latter, which may save generalization until the end of training hopefully.

However, it is hard to avoid undesired memorization from the beginning of training, since two categories can be \emph{relative to each other} and cannot be \emph{distinguished without sufficient training}.
Consider \emph{sample selection}, a correction for noisy labels where small-loss data are regarded as correct, and deep networks are trained only on selected data \citep{jiang2017mentornet,han2018coteaching,yu2019coplus}.
The losses could be \emph{enough informative after enough epochs}, but then many mislabeled data have already been memorized many times.
Consider \emph{backward correction}, where the \emph{surrogate loss} is corrected according to the label corruption process, and deep networks are trained based on this corrected loss \citep{natarajan2013learning,patrini2017making}.
The corrected loss is \emph{not necessarily a non-negative loss}, and it might go fairly negative on a training data, which signifies this data being memorized too much.
Thus, these learning methods equipped with corrections for noisy labels still suffer from undesired memorization which in turn leads to overfitting.

To relieve this issue of overfitting due to undesired memorization, we propose \emph{stochastic integrated gradient underweighted ascent}~(SIGUA).
Specifically, SIGUA belongs to stochastic optimization and can be integrated into \emph{stochastic gradient descent}~(SGD) or its variants \citep[e.g.,][]{robbins51ams,kingma2014adam}.
SIGUA works in each mini-batch: it implements SGD on good data as usual, and if there are any bad data, it implements \emph{stochastic gradient ascent}~(SGA) on bad data with a \emph{reduced learning rate}.
It is a \emph{versatile} approach, where data goodness or badness is w.r.t.\ desired or undesired memorization arose in the \emph{base} learning method.
For instance, the good-data condition for sample selection is that the loss of the deep network being trained on a data is \emph{relatively small} within the mini-batch;
that for backward correction is the corrected loss on a data is \emph{still positive}.
We can see that a good-data condition can select desired memorization, and then a bad-data condition should be designed accordingly, so that it can select undesired memorization to be relieved by SGA.

SIGUA can be justified as follows.
In machine learning, it is known that optimization shares the goal with generalization in the beginning of training, and we suffer underfitting if optimization is not well done.
However, when optimization is well done, its goal will diverge from the goal of generalization, and we suffer overfitting if optimization is too much done.
That is why we add regularizations \citep{goodfellow2016deep}, including but not limited to the powerful \emph{early stopping} \citep{morgan1990earlystop}.
Nevertheless, early stopping might not be a good choice, due to the possibility of \emph{epoch-wise double descent phenomena} occurred in training deep networks \citep{nakkiran2020ddd}.
Hence, we should trade off optimization for generalization without early stopping.
Technically, SIGUA is a specially designed regularization by \emph{pulling optimization back for generalization} when their goals conflict with each other.
A key difference between SIGUA and \emph{parameter shrinkage} like weight decay is that SIGUA pulls optimization back on some data but parameter shrinkage does the same on all data.

Furthermore, it is of vital importance to distinguish desired and undesired memorization in the presence of label noise.
Although deep networks are excellent \emph{function approximators}, they are only good at continuous functions or at least functions without \emph{jump discontinuity}.
However, memorizing mislabeled data asks the deep network being trained to exhibit a certain jump discontinuity, and thus a mislabeled data consumes notably more \emph{model capacity} than a regular data.
In terms of deep learning theory, deep networks have high \emph{adaptivity} to \emph{spatial inhomogeneity} of target function \emph{smoothness}, but a mislabeled data consumes notably more such adaptivity \citep{suzuki2019adaptivity}.
Therefore, if the deep network being trained can forget a mislabeled data, an essential amount of model capacity can be returned, which may be properly consumed later.
In this sense, philosophically, SIGUA demonstrates that \emph{forgetting undesired memorization can reinforce desired memorization}, which provides a novel viewpoint on the \emph{inductive bias of neural networks}.

\section{A Prototype of SIGUA}
\label{sec:sigua-prototype}

Let $\cX$ and $\cY$ be the input and output domains.
Consider a $k$-class classification problem, $\cY=\{1,\ldots,k\}$.
Let $(x,y)$ be the random variable pair of interest, and $p(x,y)$ be the \emph{underlying joint density} from which test data will be sampled.
In \emph{learning with noisy labels}, the training data are all sampled from a \emph{corrupted} joint density $p(x,\ytil)$ rather than $p(x,y)$, where $\ytil$ denotes the random variable of the noisy label, $p(x)$ remains the same and $p(y\mid x)$ is corrupted into $p(\ytil\mid x)$ \citep[cf.][]{natarajan2013learning,patrini2017making}:
\vspace{-1ex}%
\begin{align*}
\cS = \{(x_i,\ytil_i)\}_{i=1}^n \stackrel{\rm i.i.d.}{\sim} p(x,\ytil) = p(\ytil\mid x)p(x),
\end{align*}
where $n$ denotes the \emph{sample size} or the number of training data.
We do not use bold $x$ because $\cX$ does not necessarily belong to a vector space and $x$ is not necessarily a vector.

Let $f:\cX\to\bbR^k$ be the classifier to be trained, specifically, the \emph{score function}.
Let $\ell:\bbR^k\times\cY\to\bbR_+$ be the \emph{surrogate loss function} for $k$-class classification, e.g., \emph{softmax cross-entropy loss}.
The \emph{classification risk} of $f$ is defined as
\begin{align}
\label{eq:risk}%
R(f) = \bbE_{p(x,y)}[\ell(f(x),y)],
\end{align}
where $\bbE_{p(x,y)}$ denotes the expectation over $p(x,y)$.
If it is supervised learning where $\cS$ is drawn from $p(x,y)$, we can approximate Eq.~\eqref{eq:risk} by
\begin{align}
\label{eq:risk-emp}%
\textstyle
\Rhat(f) = \frac{1}{n}\sum_{i=1}^n \ell(f(x_i),y_i),
\end{align}
which is the \emph{empirical risk} and the objective of supervised classification before adding regularizations \citep{vapnik98SLT,goodfellow2016deep}.
The empirical risk in Eq.~\eqref{eq:risk-emp} is an unbiased estimator of the risk in Eq.~\eqref{eq:risk}, and hence the minimizer of $\Rhat(f)$ converges to the minimizer of $R(f)$ as $n$ goes to infinity \citep{vapnik98SLT}.

However, in learning with noisy labels, we cannot replace the risk with the empirical risk as $\cS$ is actually drawn from $p(x,\ytil)$.
We need some correction for noisy labels in order to approximately minimize the risk $R(f)$.
Besides certain general-purpose regularizations that might work here such as \emph{virtual adversarial training} \citep{miyato2019virtual}, there are three mainstreams---\emph{sample selection} \citep[e.g.,][]{han2018coteaching}, \emph{label correction} \citep[e.g.,][]{ma2018dimensionality}, as well as \emph{loss correction} \citep[e.g.,][]{patrini2017making}:
\vspace{-1ex}%
\begin{itemize}
    \item the first approach tries to select data with correct labels, while the second approach tries to recover correct labels for all data, so that both of them push the distribution of selected/corrected data towards $p(x,y)$;
    \item other than data manipulation, the third approach manipulates the loss, so that minimizing the expectation of the original loss over $p(x,y)$ can be rewritten into minimizing the expectation of the corrected loss over $p(x,\ytil)$.
\end{itemize}
\vspace{-1ex}%
For now, let us omit the technical details, and assume that we have a base learning method that is implemented as an algorithm with a \emph{forward pass} and a \emph{backward pass} given a mini-batch.
The forward pass returns loss values for data in this mini-batch by feeding the data through $f$, and then the backward pass returns the gradient of the average loss by propagating the average loss through $f$.
With this algorithmic abstraction, we can present SIGUA at a high level.

\paragraph{Algorithm design.}

\begin{algorithm}[t]
    \caption{SIGUA-prototype (in a mini-batch).}
    \label{alg:sigua}
    \begin{algorithmic}
        \REQUIRE base learning algorithm $\fB$, optimizer $\fO$,\\
        mini-batch $\Smb=\{(x_i,\ytil_i)\}_{i=1}^\nmb$ of batch size $\nmb$,\\
        current model $f_\theta$ where $\theta$ holds the parameters of $f$,\\
        good- and bad-data conditions $\Cg$ and $\Cb$ for $\fB$,\\
        underweight parameter $\gamma$ such that $0\le\gamma\le1$
    \end{algorithmic}
    \begin{algorithmic}[1]
        \STATE $\{\ell_i\}_{i=1}^\nmb \gets \fB$.forward($f_\theta$, $\Smb$) \hfill \# forward pass
        \STATE $\lmb \gets 0$ \hfill \# initialize loss accumulator
        \FOR{$i=1,\ldots,\nmb$}
        \IF{$\Cg(x_i,\ytil_i)$}
        \STATE $\lmb \gets \lmb + \ell_i$ \hfill \# accumulate loss positively
        \ELSIF{$\Cb(x_i,\ytil_i)$}
        \STATE $\lmb \gets \lmb - \gamma\ell_i$ \hfill \# accumulate loss negatively
        \ENDIF \hfill \# ignore any uncertain data
        \ENDFOR
        \STATE $\lmb \gets \lmb/\nmb$ \hfill \# average accumulated loss
        \STATE $\nabla_\theta \gets \fB$.backward($f_\theta$, $\lmb$) \hfill \# backward pass
        \STATE $\fO$.step($\nabla_\theta$) \hfill \# update model
    \end{algorithmic}
\end{algorithm}

A prototype of \emph{stochastic integrated gradient underweighted ascent}~(SIGUA) is given in Algorithm~\ref{alg:sigua}.
Since it is only a prototype, it serves as a versatile approach where the meanings of different steps depend on the base learning algorithm $\fB$ (i.e., Lines~1, 4, 6 and 11).

More specifically, $\Cg,\Cb:\cX\times\cY\to\{0,1\}$ are functions mapping $(x_i,\ytil_i)$ to either true or false:
\vspace{-1ex}%
\begin{itemize}
    \item if it is a good data, $\Cg$/$\Cb$ returns true/false;
    \item if it is a bad data, $\Cg$/$\Cb$ returns false/true;
    \item otherwise, $\Cg$ and $\Cb$ both return false.
\end{itemize}
\vspace{-1ex}%
The last option is of conceptual importance, which allows $\Cg$ and $\Cb$ not to cover all data, but to leave some data that we are uncertain about to be regarded as \emph{neither good nor bad}.
We have assumed $\Cg$ and $\Cb$ are functions of $x$ and $\ytil$ for simplicity; they may also require other data in $\Smb$ or other information about $\fB$ and $\fO$ in reality.

Algorithm~\ref{alg:sigua} runs as follows.
Given the mini-batch $\Smb$, the forward pass of $\fB$ is called in Line~1.
Then the loss values are manipulated in Lines~2--10 where they are reduced to a scalar ready for \emph{backpropagation}.
Before the for loop, the \emph{loss accumulator} $\lmb$ is initialized in Line~2. Subsequently,
\vspace{-1ex}%
\begin{itemize}
    \item in Line~5, $\ell_i$ is added to $\lmb$ if $(x_i,\ytil_i)$ meets $\Cg$, which will result in \emph{gradient descent} by $\fO$ in Line~12;
    \item in Line~7, $\ell_i$ is underweighted by $\gamma$ and subtracted from $\lmb$ if $(x_i,\ytil_i)$ meets $\Cb$, which will lead to \emph{gradient underweighted ascent} by $\fO$ in Line~12;
    \item otherwise, no branch is executed, so that $\ell_i$ is ignored in $\lmb$, which will cause \emph{stop gradient} by $\fO$ in Line~12.
\end{itemize}
\vspace{-1ex}%
After the for loop, the accumulated loss $\lmb$ is divide by the batch size $\nmb$ in Line~10 to make the average loss.
Finally, the backward pass of $\fB$ is called in Line~11, and the optimizer $\fO$ comes to update the current model $f_\theta$ in Line~12.

In order to integrate gradient ascent within an optimizer $\fO$ carrying out gradient descent, a loss accumulator suffices, and it is more efficient than a \emph{gradient accumulator}.
There is no difference between negating losses and negating gradients, while $\gamma$ has the same effects in reducing losses and reducing the learning rate inside $\fO$.
If using deep learning framework based on \emph{dynamic computational graph} \citep{tokui2015chainer} such as PyTorch and TensorFlow eager execution, we can modify losses \emph{in-place} instead of accumulate them.
In practice, we can also get rid of the for loop using a \emph{computationally more efficient} implementation of Algorithm~\ref{alg:sigua}.
Suppose the forward pass of $\fB$ returns $\blmb\in\bbR^\nmb$, i.e., a vector but not a set of loss values, and $\Cg$ and $\Cb$ directly map $\Smb$ to $\{0,1\}^\nmb$, i.e., two vectors of good- and bad-data masks.
Then, Lines~2--10 may be replaced with a single line:
\vspace{-1ex}%
\begin{align}
\label{eq:sigua}%
\lmb \gets \left( \blmb^\T(\Cg(\Smb)-\gamma\Cb(\Smb)) \right)/\nmb,
\end{align}
where $\T$ denotes the transpose, and $\Cg(\Smb)-\gamma\Cb(\Smb)$ is a vector whose entries take $1$ for good data, $0$ for uncertain data, and $-\gamma$ for bad data.
Eq.~\eqref{eq:sigua} includes everything about SIGUA, and thus we will refer to either Algorithm~\ref{alg:sigua} or Eq.~\eqref{eq:sigua} as SIGUA, interchangeably.

Last but not least, notice that SIGUA is extremely general.
SIGUA becomes \emph{standard training}, if $\Cg$ always returns true.
It becomes \emph{training on good data only}, if $\Cb$ always returns false; we name it StopGrad because SIGUA is also named after how we handle non-good data.
Moreover, the hyperparameter $\gamma$ controls the strength of gradient ascent: when $\gamma=0$, it becomes StopGrad again; when $\gamma=1$, the bad data would be erased by gradient full ascent instead of gradient underweighted ascent.
Consequently, $\gamma$ should be carefully tuned on validation data in practice.

\paragraph{Motivations of design.}
The idea of SIGUA is motivated in the introduction, and its specific algorithm design is motivated here.
At least four questions can be raised:
\vspace{-1ex}%
\begin{itemize}[leftmargin=2em]
    \item[Q1] What $\fB$ can be used and what are its $\Cg$ and $\Cb$?
    \item[Q2] What is the technical implication of gradient ascent?
    \item[Q3] Why gradient ascent is necessary for training $f_\theta$?
    \item[Q4] Why underweight is necessary for gradient ascent?
\end{itemize}
\vspace{-1ex}%
Among these questions, Q1 is most complicated---we will devote the entire Section~\ref{sec:sigua-realization} for answering it; the other three questions are answered below one by one.

The technical implication of gradient ascent depends on $k$ and $\ell$.
Consider binary classification, and assume $\ell:\bbR\times\cY\to\bbR_+$ satisfies a \emph{symmetric condition} \citep{christo2014nips,niu2016nips}: $\ell(t,+1)+\ell(t,-1)=\const$, for example \emph{ramp loss} and \emph{sigmoid loss} \citep[cf.][]{kiryo2017positive}.
Then, we can obtain that
\begin{align*}
-\nabla_\theta\ell(f_\theta(x_i),\ytil_i)=\nabla_\theta\ell(f_\theta(x_i),-\ytil_i),
\end{align*}
which indicates that ascent along such a gradient is equivalent to descent along another gradient where $\ytil_i$ is flipped.
For binary classification, $-\ytil_i$ must be correct if $\ytil_i$ is incorrect for $x_i$, which implies SIGUA is exactly same as label correction.
When $\ell$ does not satisfy that symmetric condition, they become different but still conceptually similar.

That being said, for multi-class classification where $k\ge3$, we cannot know which class is correct if $\ytil_i$ is incorrect for $x_i$.
Hence, we let the model \emph{forget the wrong information} that $x_i$ is from class $\ytil_i$.
This forgetting behavior can return some capacity back to the model, and later the model may use this capacity in a better way.
This forgetting behavior is the key of SIGUA and what we meant by \emph{forgetting may make learning with noisy labels more robust} in the title.

\begin{figure}[t]
    \begin{minipage}[c]{0.05\columnwidth}~\end{minipage}%
    \begin{minipage}[c]{0.475\columnwidth}\centering\small | StopGrad | \end{minipage}%
    \begin{minipage}[c]{0.475\columnwidth}\centering\small | SIGUA | \end{minipage}\\
    \begin{minipage}[c]{0.05\columnwidth}\centering\small \rotatebox[origin=c]{270}{| Wide net |} \end{minipage}%
    \begin{minipage}[c]{0.95\columnwidth}
        \includegraphics[width=0.5\textwidth]{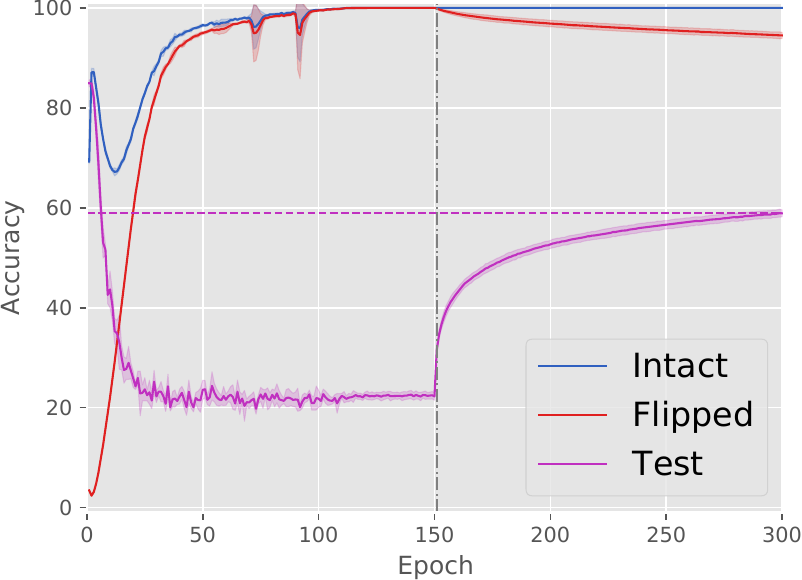}%
        \includegraphics[width=0.5\textwidth]{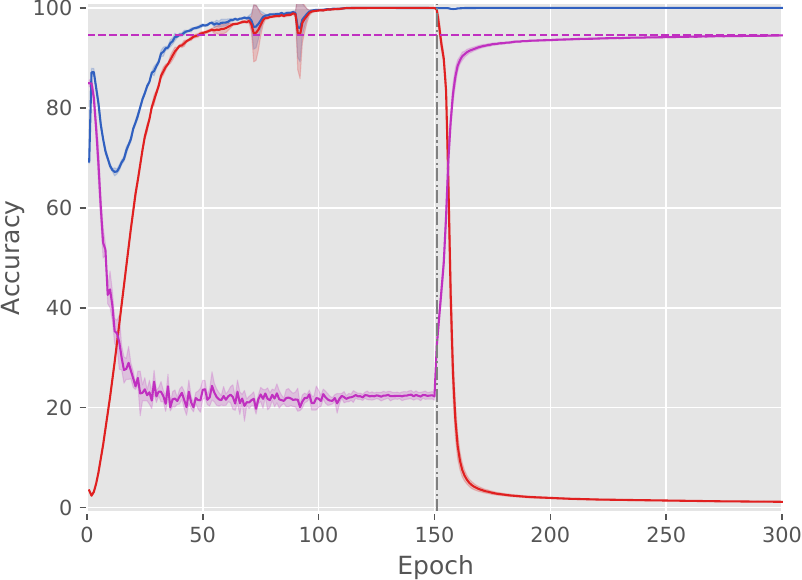}%
    \end{minipage}\\
    \begin{minipage}[c]{0.05\columnwidth}\centering\small \rotatebox[origin=c]{270}{| Deep net |} \end{minipage}%
    \begin{minipage}[c]{0.95\columnwidth}
        \includegraphics[width=0.5\textwidth]{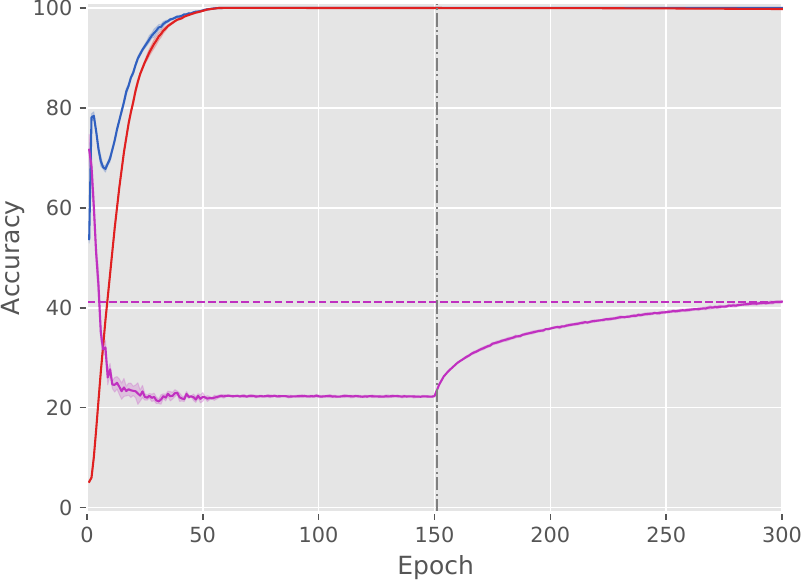}%
        \includegraphics[width=0.5\textwidth]{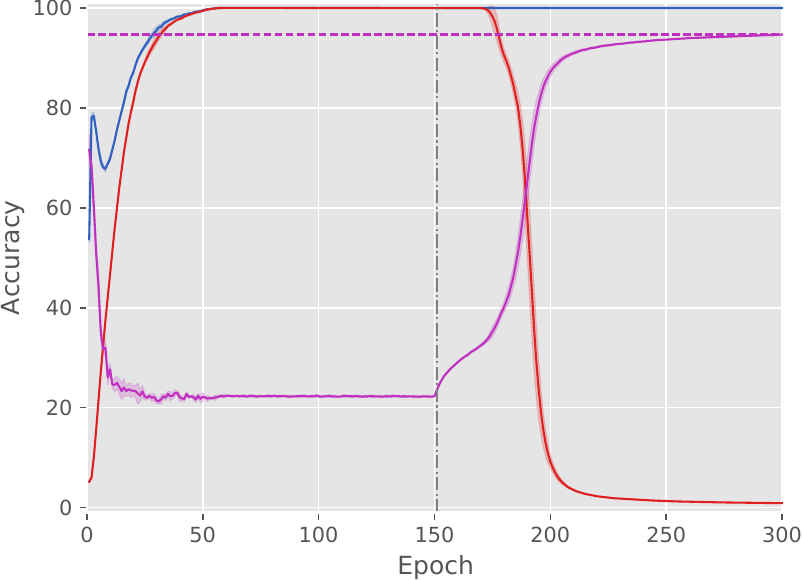}%
    \end{minipage}
    \vspace{-1em}%
    \caption{StopGrad vs.~SIGUA on noisy MNIST.}
    \label{fig:stopgrad-vs-sigua}
    \vspace{-1em}%
\end{figure}

Concerning why gradient ascent is necessary or why StopGrad is inadequate, we answer it with empirical evidence.
We take the MNIST benchmark dataset and add 80\% \emph{symmetric label noise}: for each $x_i$, $\pr(\ytil_i=y_i)=0.2$ and for all $y\neq y_i$, $\pr(\ytil_i=y)=0.8/9$.
Two neural networks are considered in this experiment:
\vspace{-1ex}%
\begin{itemize}
    \item a wide network with an architecture 784-[Lin(10k)-BN-ReLU]-[Lin(100)-BN-ReLU]-Lin(10) which has \emph{8,871k} parameters coming from 5 trainable layers;
    \item a deep network with an architecture 784-[Lin(500)-BN-ReLU]*5-Lin(10) which has \emph{1,405k} parameters coming from 11 trainable layers.
\end{itemize}
\vspace{-1ex}%
Note that MNIST contains 60k training data, where \emph{12k is intact} and \emph{48k is flipped}, and thus the two neural networks are clearly over-parameterized.
The base learning method $\fB$ is standard training that treats $\Smb$ as drawn from $p(x,y)$ where $\nmb$ is 1024.
The optimizer $\fO$ is the popular \emph{momentum SGD}, where the learning rate and momentum are fixed to 0.01 and 0.9---there is no learning rate decay, otherwise StopGrad may be inadequate due to learning rate decay.

Since this experiment is for illustrating our motivation, we design $\Cg$ and $\Cb$ using the true labels.
The number of epochs is 300 and the shift is between 150 and 151: before the shift, we let $\Cg(x_i,\ytil_i)=1$, and then StopGrad and SIGUA would fit all training data; after the shift, we let
\begin{align*}
\Cg(x_i,\ytil_i)=\bbI(\ytil_i=y_i),\quad
\Cb(x_i,\ytil_i)=\bbI(\ytil_i\neq y_i),
\end{align*}
where $\bbI$ denotes the \emph{indicator function} and it tests a conditional expression, and then they would fit only intact data.
We let $\gamma=0$ for StopGrad and $\gamma=0.001$ for SIGUA, and as a result StopGrad would ignore flipped data but SIGUA would counter-fit flipped data during epochs 151--300.
We repeat this random label flipping and training 5 times.

The experimental results are shown in Figure~\ref{fig:stopgrad-vs-sigua}, where the means with standard deviations of the accuracy curves are plotted.
In Figure~\ref{fig:stopgrad-vs-sigua}, blue means training accuracy on intact data, red means that on flipped data, and purple means test accuracy on label-noise-free test data.
We can see that
\vspace{-1ex}%
\begin{itemize}
    \item the wide and deep networks can both memorize all data, while \emph{perfect memorization} occurred around epoch 120 for the wide one and epoch 60 for the deep one, namely the deep memorized faster than the wide;
    \item in fact the wide not only memorized slower but also forgot faster, implying that its model capacity is lower than the deep, even though it has 5 times more parameters;
    \item at the end of training, StopGrad made the wide and deep networks forget 5\% and 0\% flipped data, while their test accuracy was improved from 23\% to 59\% and 41\%;%
    \footnote{This is because the softmax cross-entropy loss on intact data can still further approach to zero after perfect memorization.}
    \item SIGUA made both of them forget 99\% flipped data and their test accuracy was improved from 23\% to 95\%.
\end{itemize}
\vspace{-1ex}%
Note that MNIST contains 10 classes, so that the accuracy on flipped data lower than 10\% means that perfect memorization has already been erased.
Indeed, SIGUA made the accuracy on flipped data lower than 1\%, which means that the model memorized a fact that the labels of those flipped data are flipped.
In other words, SIGUA achieved \emph{learning with complimentary labels} on those flipped data implicitly \citep{ishida2017complementary,yu2018complementary,ishida2019complementary,feng2020complementary,chou2020complementary}.

Lastly, the necessity of the underweight parameter $\gamma$ is for the \emph{stability of optimization}.
Without underweight for gradient ascent, training may be completely destroyed, if $\fO$ is sophisticated and uses adaptive learning rates for different parameters such as \emph{Adam} \citep{kingma2014adam}.

\section{Two Realizations of SIGUA}
\label{sec:sigua-realization}

In this section, we explain what $\fB$ can be used in SIGUA.
We employ SIGUA to robustify \emph{self-teaching} that belongs to the sample-selection approach and \emph{backward correction} that belongs to the loss-correction approach.
Self-teaching and backward correction are two \emph{representative} and more importantly \emph{orthogonal} methods in learning with noisy labels, which spotlights the great versatility of SIGUA.

\paragraph{SIGUA robustifies self-teaching.}
The sample-selection approach regards \emph{small-loss data} as ``correct'', and it trains the model $f_\theta$ only on selected small-loss data \citep{jiang2017mentornet,han2018coteaching,yu2019coplus}.
Self-teaching, or equivalently \emph{self-paced MentorNet} \citep{jiang2017mentornet},%
\footnote{Technically, self-paced MentorNet uses \emph{sample reweighting}, but its idea is essentially similar to self-teaching.}
is the most primitive method in this direction.
It maintains a single model, selects small-loss data as useful knowledge, and teaches this knowledge to itself.

In order to select small-loss data, a parameter of the label corruption process is needed---the \emph{noise level} $\epsilon$ measuring how many labels are corrupted.
Note that $\epsilon$ is a scalar and can be easily estimated in practice \citep[e.g.,][]{liu2016classification,patrini2017making}.
Then, the rate of data to be selected is
\begin{align}
\label{eq:rho}%
\rho(t) = 1 - \epsilon\cdot\min(t/T_k,1),
\end{align}
where $t$ is the current epoch number, and $T_k$ is a hyperparameter denoting the number of epochs for \emph{warm-up}.
This means we gradually select more and more data in the first $T_k$ epochs, and then we select a fixed amount of data from epoch $T_k$, as the losses are unreliable after $f_\theta$ is randomly initialized and they become more and more reliable during training \citep{han2018coteaching}.
Subsequently, let us denote by $\ell_i=\ell(f_\theta(x_i),\ytil_i)$ for $(x_i,\ytil_i)\in\Smb$ and define
\begin{align}
\label{eq:cond-good-small-loss}%
\textstyle
\Cg(x_i,\ytil_i) = \bbI\left( \sum_{j=1}^\nmb\bbI(\ell_i>\ell_j) \le \nmb\rho(t) \right),
\end{align}
where $\bbI(\ell_i>\ell_j)$ tests if the loss on $(x_i,\ytil_i)$ is greater than the loss on $(x_j,\ytil_j)$, and then $\sum_{j=1}^\nmb\bbI(\ell_i>\ell_j)$ counts how many data in $\Smb$ have smaller losses than $\ell_i$.
Eq.~\eqref{eq:cond-good-small-loss} is true if the count is smaller than or equal to $\nmb\rho(t)$, i.e., $(x_i,\ytil_i)$ is a small-loss data in $\Smb$.
Self-teaching can be realized by plugging Eq.~\eqref{eq:cond-good-small-loss} and $\Cb(x_i,\ytil_i)=0$ into Eq.~\eqref{eq:sigua}.

Eq.~\eqref{eq:rho} claims loss values can be enough informative after $T_k$ epochs, at which time many mislabeled data have been memorized many times.
Moreover, small-loss data are just \emph{likely to be correct but not certainly}, so that incorrect sample selection will mislead the model training which will in turn mislead the selection next time.
As a consequence, we should employ SIGUA to robustify self-teaching.
Let $\delta(t)$ be the rate of data to be forgotten, then $\Cb(x_i,\ytil_i)$ can be defined similarly as
\begin{align}
& \Cb(x_i,\ytil_i) = \neg \Cg(x_i,\ytil_i) \land \notag\\
\label{eq:cond-bad-small-loss}%
&\textstyle\qquad \bbI\left( \sum_{j=1}^\nmb\bbI(\ell_i>\ell_j) \le \nmb\rho(t)+\nmb\delta(t) \right),
\end{align}
where $\neg \Cg(x_i,\ytil_i)$ is necessary for Eq.~\eqref{eq:sigua} but not Algorithm~\ref{alg:sigua}.
We refer to this self-teaching enhanced by Eq.~\eqref{eq:cond-bad-small-loss} as SIGUA$\SL$ where SL stands for small loss.

It seems counter-intuitive to select \emph{middle-loss data} rather than \emph{large-loss data} as our bad data.
In fact, large-loss data are not memorized very well---no hope to confuse $\Cg$ in Eq.~\eqref{eq:cond-good-small-loss}, and then no need to be selected by $\Cb$ in Eq.~\eqref{eq:cond-bad-small-loss}.
On the other hand, similarly to large-loss data, middle-loss data might be mislabeled with high probability, while they are memorized relatively well.
To this end, we would like the model to slightly forget these middle-loss data, and let $\Cg$ be less confused by them.
This motivates the design of the bad-data condition $\Cb$ in Eq.~\eqref{eq:cond-bad-small-loss}.

\paragraph{SIGUA robustifies backward correction.}
On the other hand, the loss-correction approach creates a \emph{corrected loss} from $\ell$ and then trains the model $f_\theta$ based on the corrected loss \citep{patrini2017making}.
Backward correction, for binary classification \citep{natarajan2013learning} or multi-class classification \citep{patrini2017making}, is the most primitive method in this direction.
It builds $\ellb:\bbR^k\times\cY\to\bbR$ to reverse the label corruption process, and minimizes
\begin{align}
\label{eq:risk-emp-corr}%
\textstyle
\Rhatb(f) = \frac{1}{n}\sum_{i=1}^n \ellb(f(x_i),\ytil_i),
\end{align}
which is the \emph{corrected empirical risk}.

In order to reverse the label corruption process, a model of it is needed.
Note that $p(\ytil\mid x)=\sum_yp(\ytil\mid x,y)p(y\mid x)$.
A model for $p(\ytil\mid x,y)$ is called \emph{instance-dependent noise} \citep[cf.][]{menon2018idn,cheng2020idn,berthon2020idn}, which is unfortunately \emph{unidentifiable} without some extra assumption/information.
Thus, a common practice is to assume that $p(\ytil\mid x,y)=p(\ytil\mid y)$, i.e., the corruption is \emph{instance-independent} and \emph{class-conditional}. This model is called \emph{class-conditional noise}~(CCN).
Let $T\in\bbR_+^{k\times k}$ be a \emph{transition matrix} such that $[T]_{i,j}=p(\ytil=j\mid y=i)$.
This $T$ is much more difficult to estimate than $\epsilon$, which is a hot topic in learning with noisy labels and there are still many methods \citep{liu2016classification,patrini2017making,han2018masking,hendrycks2018using,xia2019anchor,yao2020dual,xia2020parts}.
Using $T$, $\ellb$ is defined as
\begin{align}
\label{eq:loss-back-corr}%
\ellb(f(x),\ytil) = [T^{-1}\bell]_{\ytil},
\end{align}
where $\bell=(\ell(f(x),1),\ldots,\ell(f(x),k))\in\bbR_+^k$, and it holds that \citep[Theorem 1]{patrini2017making}
\begin{align}
\label{eq:guarantee-back-corr}%
\bbE_\cS[\Rhatb(f)] = \bbE_{p(x,\ytil)}[\ellb(f(x),\ytil)] = R(f),
\end{align}
i.e., $\Rhatb(f)$ is an \emph{unbiased estimator} of $R(f)$ or backward correction is \emph{risk-consistent}.

Nonetheless, we should be careful of the above theoretical guarantee.
Eq.~\eqref{eq:guarantee-back-corr} is about the \emph{asymptotic} case rather than the \emph{finite-sample} case.
Note that $T^{-1}\in\bbR^{k\times k}$ though $T\in\bbR_+^{k\times k}$, and then $\ellb$ is no longer a non-negative loss.
Since we are minimizing $\ellb(f_\theta(x_i),\ytil_i)$ where $(x_i,\ytil_i)$ is from a finite sample $\cS$ and $f_\theta$ is an over-parameterized model, the loss \emph{must go negative} sooner or later whenever it \emph{could go negative} \citep{kiryo2017positive,ishida2019complementary,lu2019mitigating}.
If $\ellb(f_\theta(x_i),\ytil_i)$ is fairly negative, it signifies that $(x_i,\ytil_i)$ has been memorized too much and it suggests that the optimizer $\fO$ should focus on the data whose losses are still positive.
Consequently, we should employ SIGUA to robustify backward correction.

More specifically, we have two choices, as there is no specific good-data condition yet: one focuses on $\ellb(f(x),\ytil)$, and the other focuses on $T^{-1}\bell$ as a whole.
We adopt the latter one, since requiring $\ellb(f(x),\ytil)\ge0$ may be too strict and aggressive.
Taking a closer look at the derivation of backward correction, we can see that for any $x$,
\begin{align}
\label{eq:guarantee-x-back-corr}%
\bpyt^\T\bellb = \bpy^\T\bell,
\end{align}
where $\bpy=(p(y=1\mid x),\ldots,p(y=k\mid x))$, $\bpyt$ is as $\bpy$, and $\bellb=T^{-1}\bell$.
In Eq.~\eqref{eq:guarantee-x-back-corr}, the right-hand side is always non-negative, and so should be the left-hand side.
However, $\bpyt$ is unknown to us, and thus we replace it with the uninformative uniform distribution.
Finally, denote by $\bone$ the all-one vector in $\bbR^k$, and then the good- and bad-data conditions can be defined as
\begin{align}
\label{eq:cond-good-back-corr}%
\Cg(x_i,\ytil_i) &= \bbI\left( \bone^\T\bellb\ge0 \right),\\
\label{eq:cond-bad-back-corr}%
\Cb(x_i,\ytil_i) &= \neg \Cg(x_i,\ytil_i).
\end{align}
We refer to this backward correction enhanced by Eq.~\eqref{eq:cond-good-back-corr} and Eq.~\eqref{eq:cond-bad-back-corr} as SIGUA$\BC$.

\section{Experiments}
\label{sec:experiments}

\begin{figure*}[t]
    \centering
    \begin{minipage}[c]{0.05\columnwidth}~\end{minipage}%
    \begin{minipage}[c]{0.3\textwidth}\centering\small || Symmetry-20\% || \end{minipage}%
    \begin{minipage}[c]{0.3\textwidth}\centering\small || Symmetry-50\% || \end{minipage}%
    \begin{minipage}[c]{0.3\textwidth}\centering\small || Pair-45\% || \end{minipage}\\
    \begin{minipage}[c]{0.05\columnwidth}\centering\small \rotatebox[origin=c]{270}{|| MNIST ||} \end{minipage}%
    \begin{minipage}[c]{0.9\textwidth}
        \includegraphics[width=0.33\textwidth]{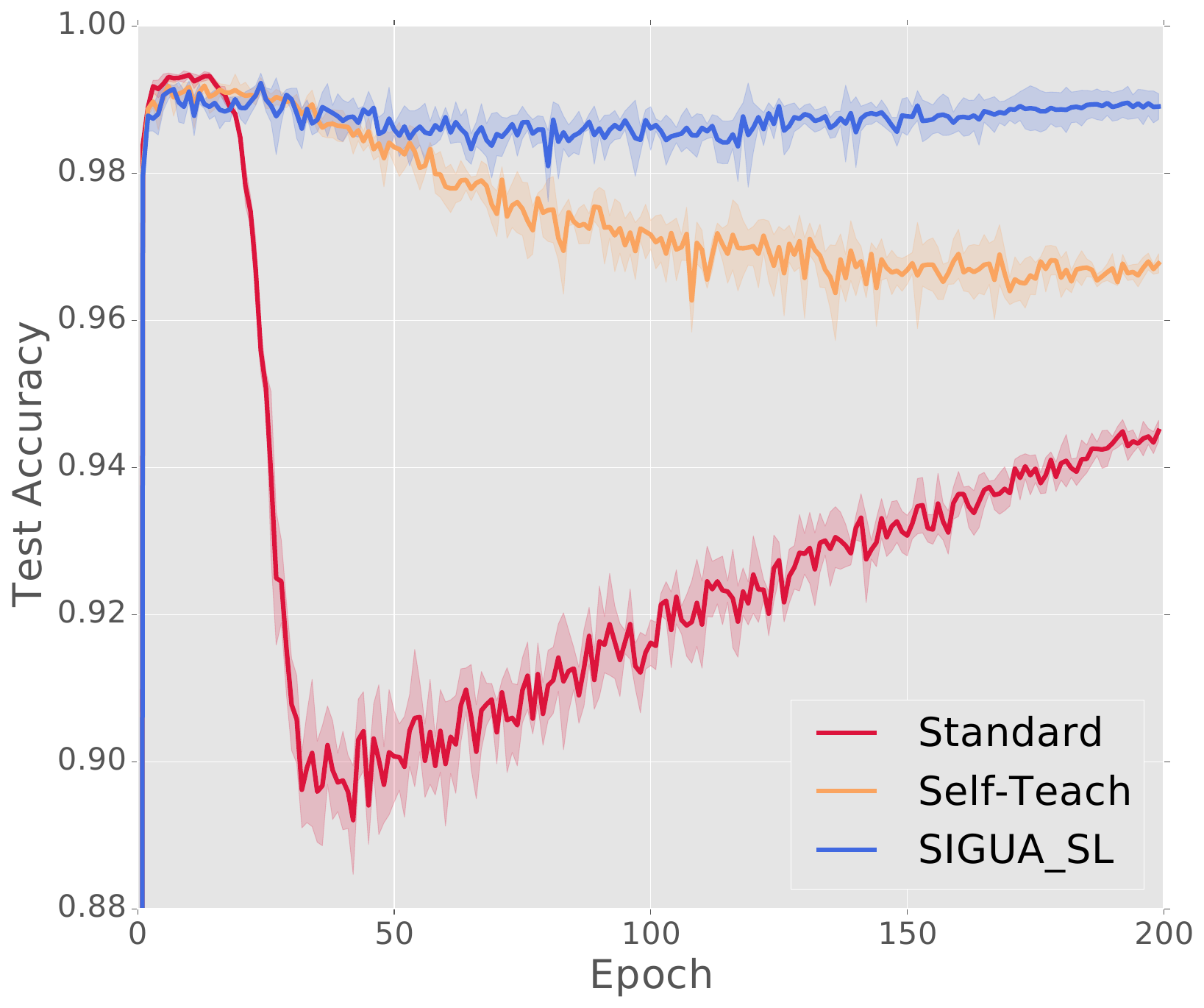}%
        \includegraphics[width=0.33\textwidth]{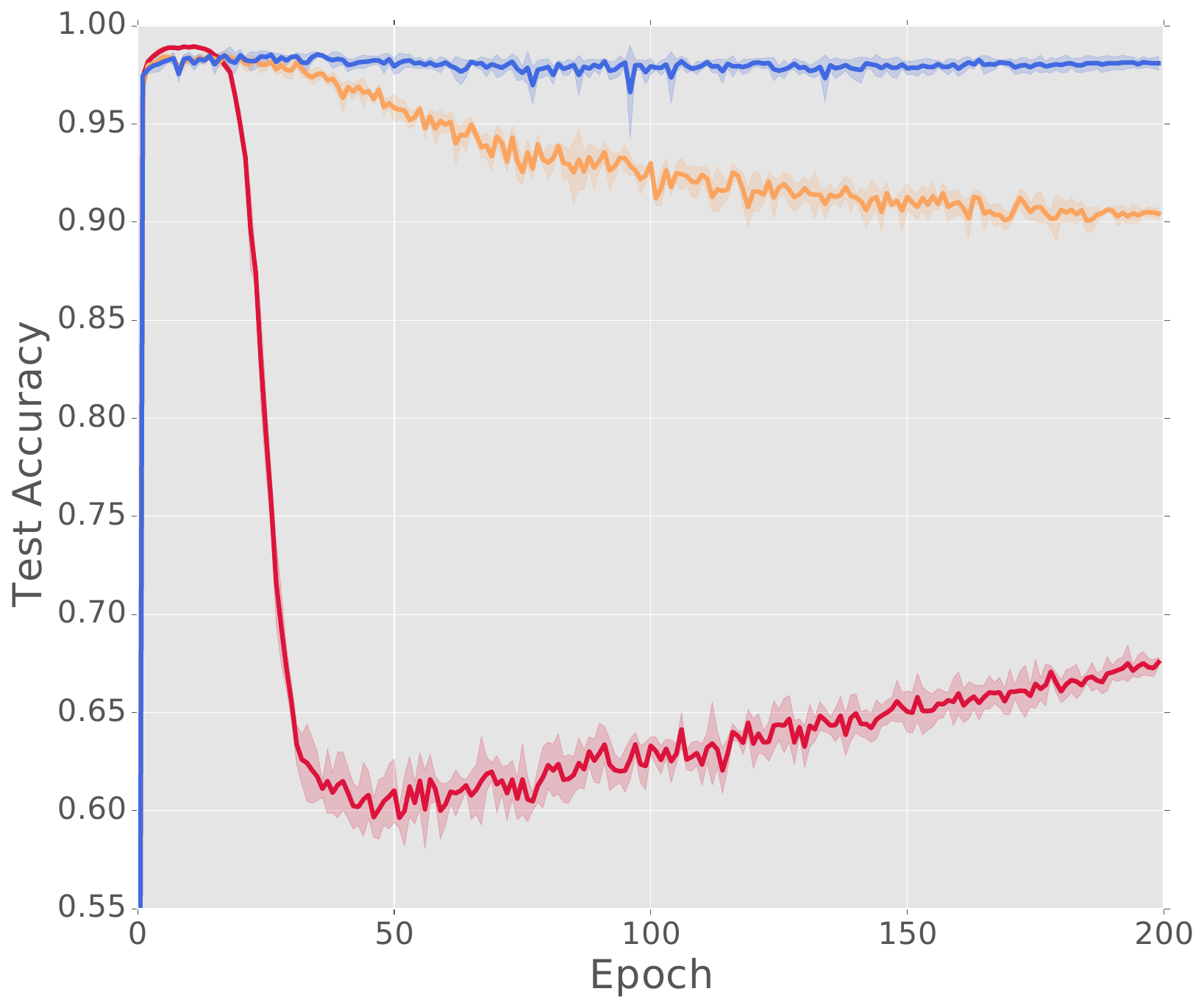}%
        \includegraphics[width=0.33\textwidth]{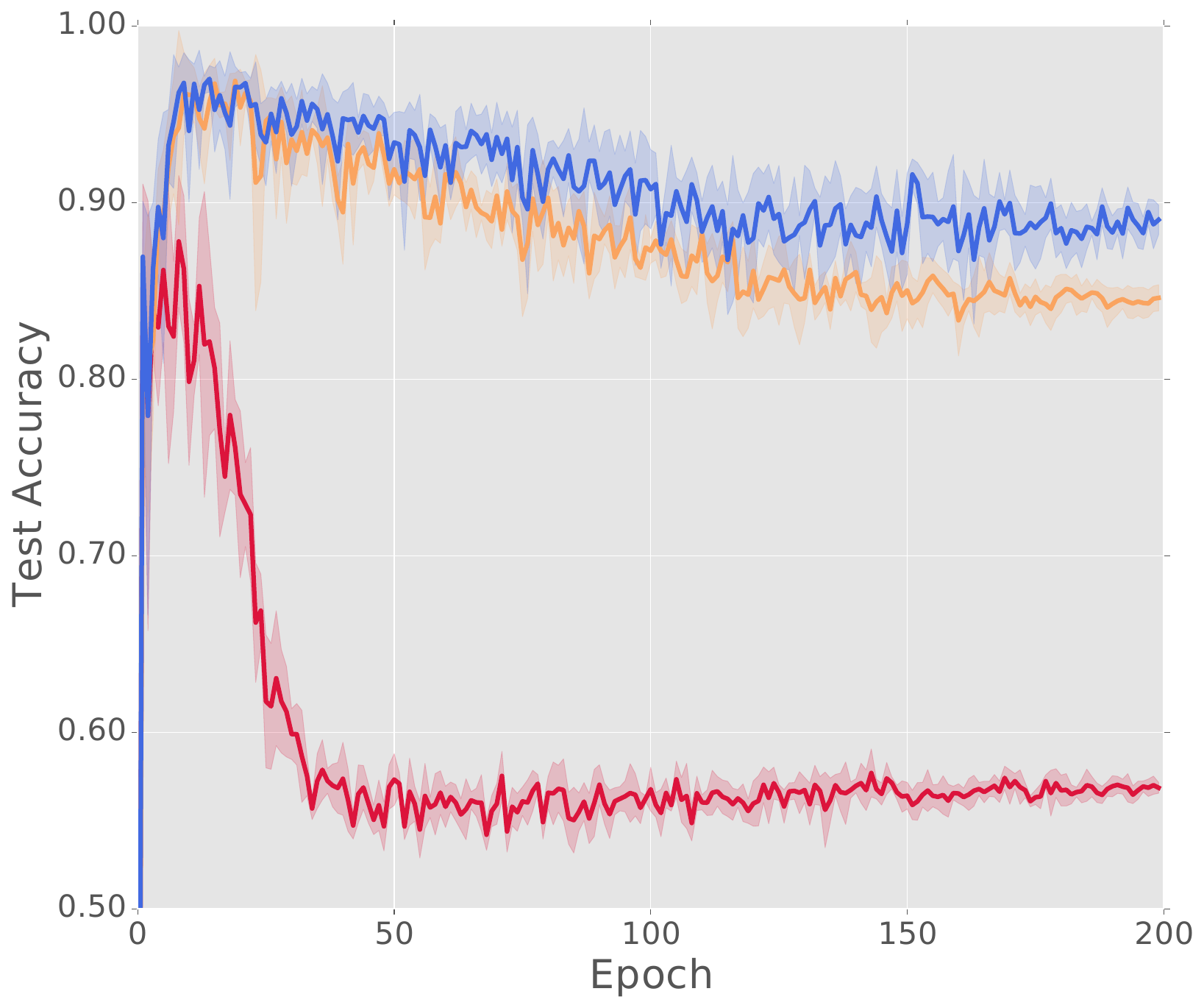}%
    \end{minipage}\\
    \begin{minipage}[c]{0.05\columnwidth}\centering\small \rotatebox[origin=c]{270}{|| CIFAR-10 ||} \end{minipage}%
    \begin{minipage}[c]{0.9\textwidth}
        \includegraphics[width=0.33\textwidth]{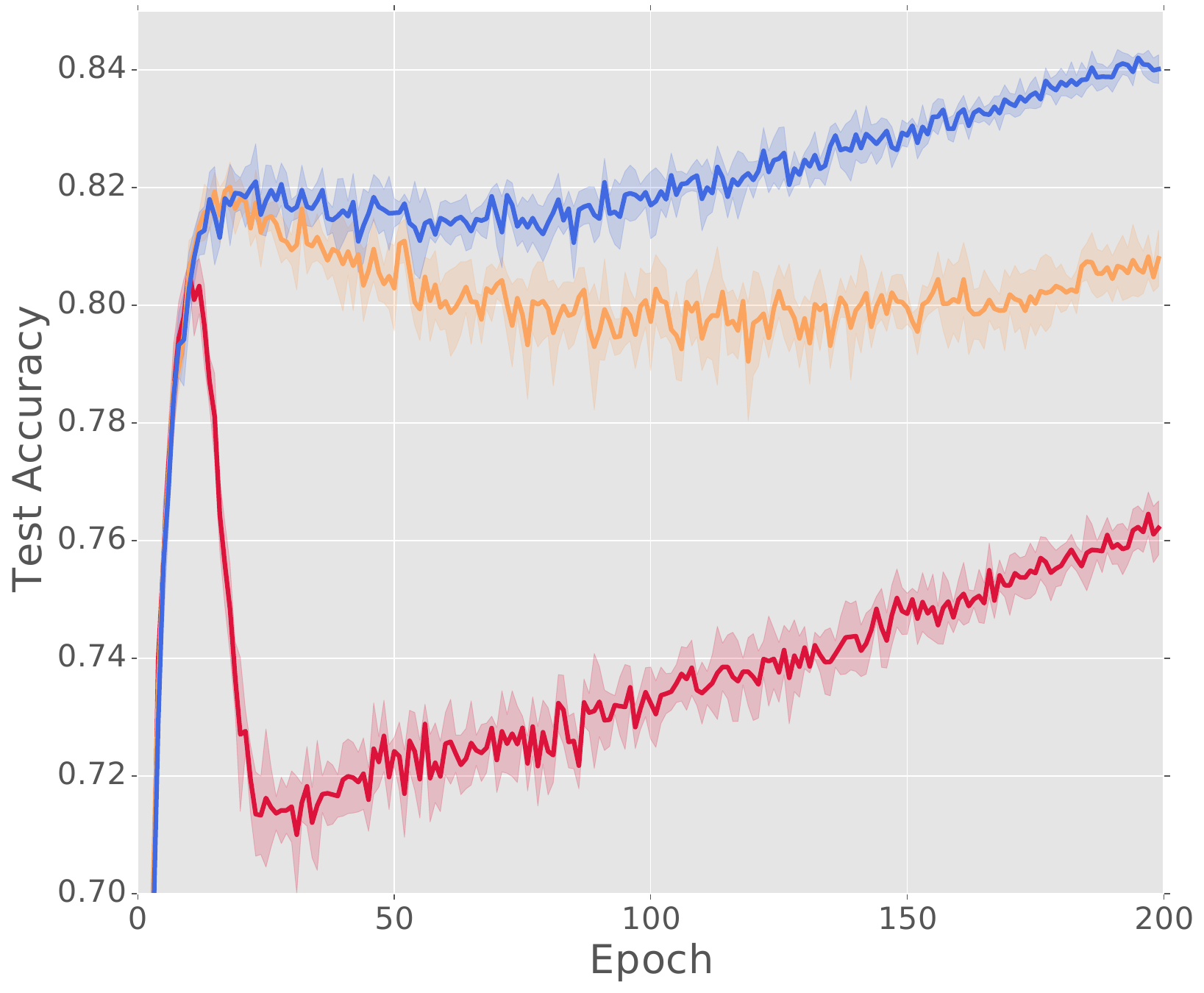}%
        \includegraphics[width=0.33\textwidth]{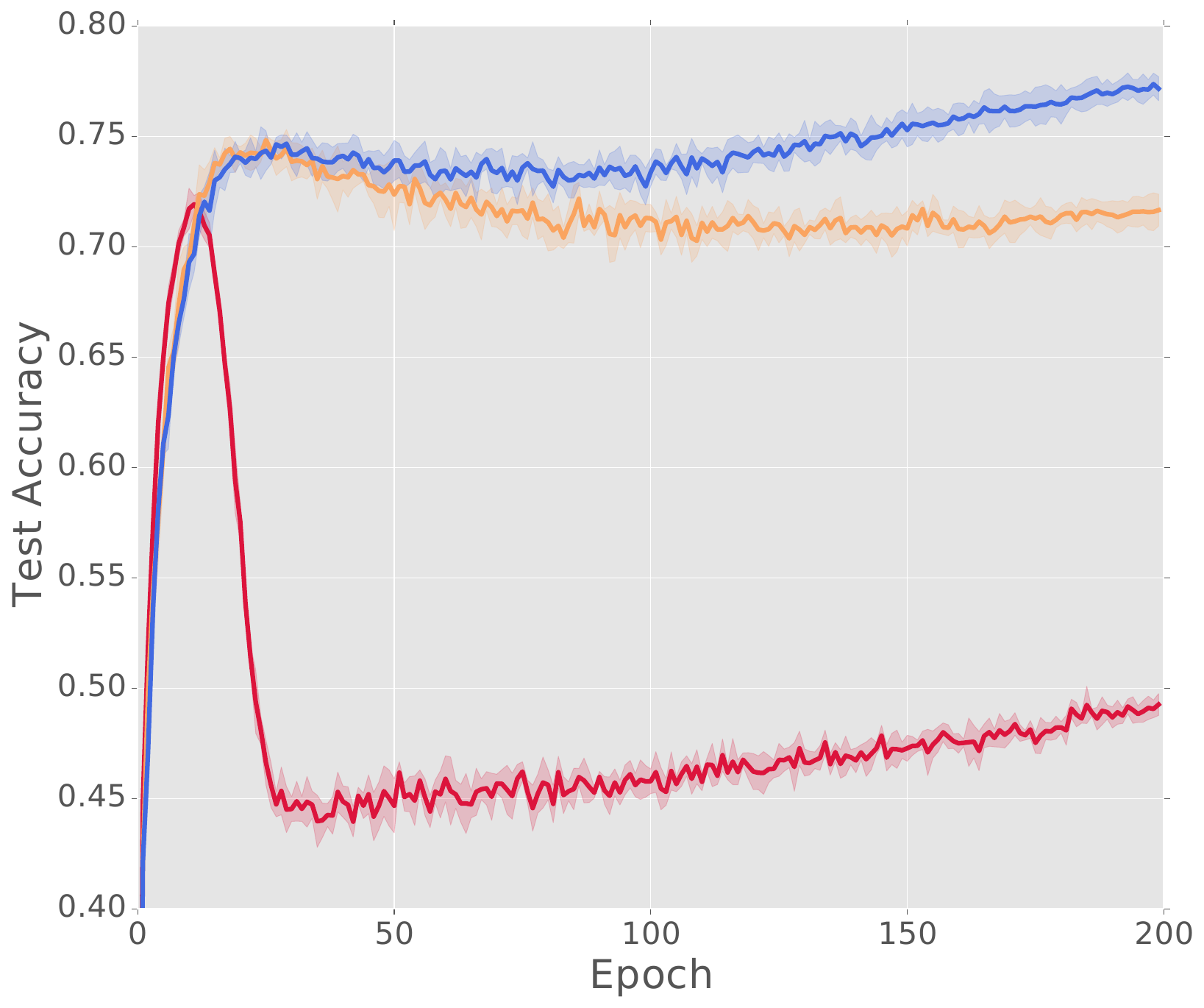}%
        \includegraphics[width=0.33\textwidth]{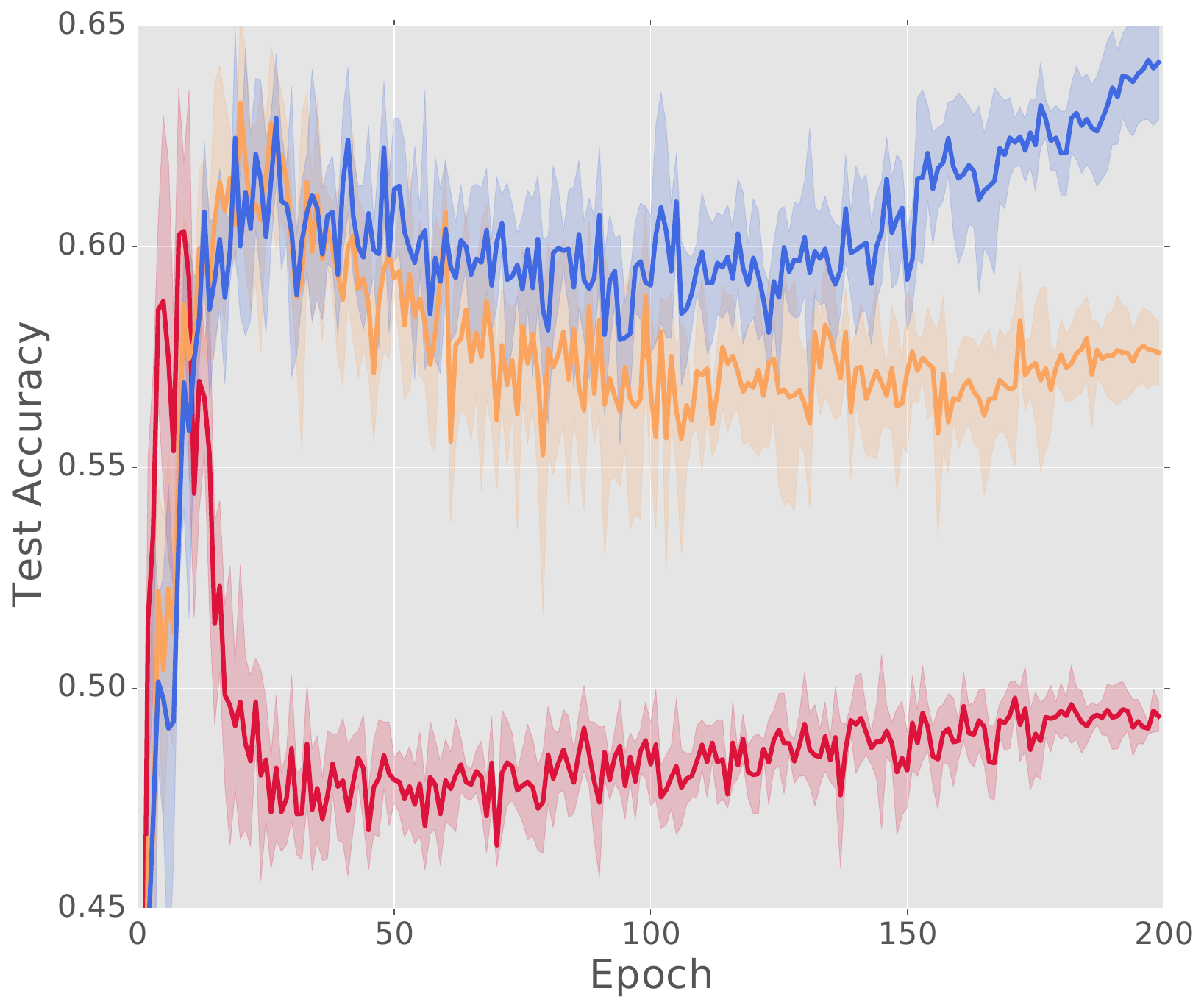}%
    \end{minipage}\\
    \vspace{-1ex}%
    \caption{Accuracy curves of training deep networks using the three learning methods in SET1.}
    \label{fig:small-loss-mnist-cifar10}
    \vspace{-1ex}%
\end{figure*}

\begin{figure*}
    \centering
    \begin{minipage}[c]{0.05\columnwidth}~\end{minipage}%
    \begin{minipage}[c]{0.3\textwidth}\centering\small || Symmetry-20\% || \end{minipage}%
    \begin{minipage}[c]{0.3\textwidth}\centering\small || Symmetry-50\% || \end{minipage}%
    \begin{minipage}[c]{0.3\textwidth}\centering\small || Pair-45\% || \end{minipage}\\
    \begin{minipage}[c]{0.05\columnwidth}\centering\small \rotatebox[origin=c]{270}{|| MNIST ||} \end{minipage}%
    \begin{minipage}[c]{0.9\textwidth}
        \includegraphics[width=0.33\textwidth]{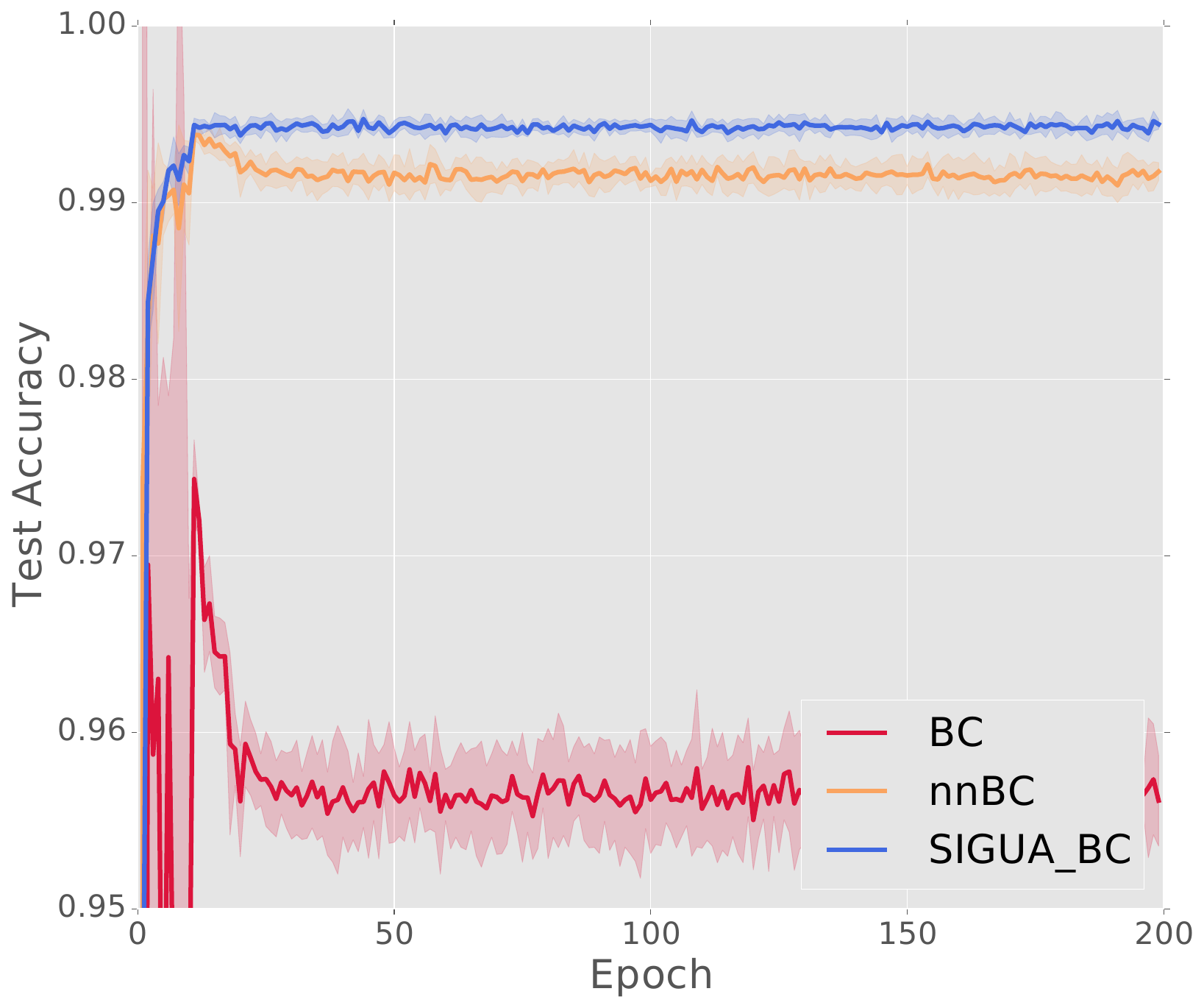}%
        \includegraphics[width=0.33\textwidth]{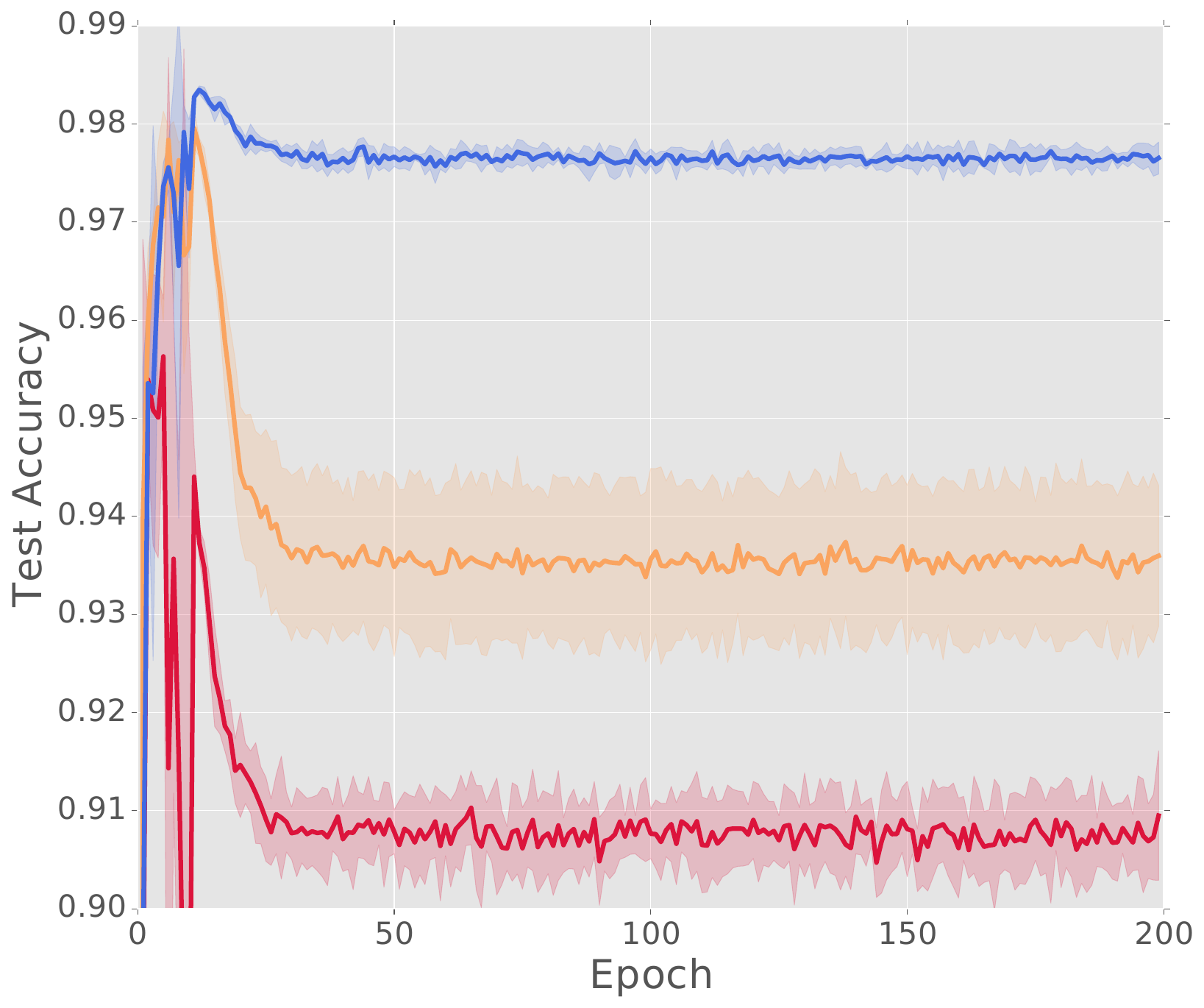}%
        \includegraphics[width=0.33\textwidth]{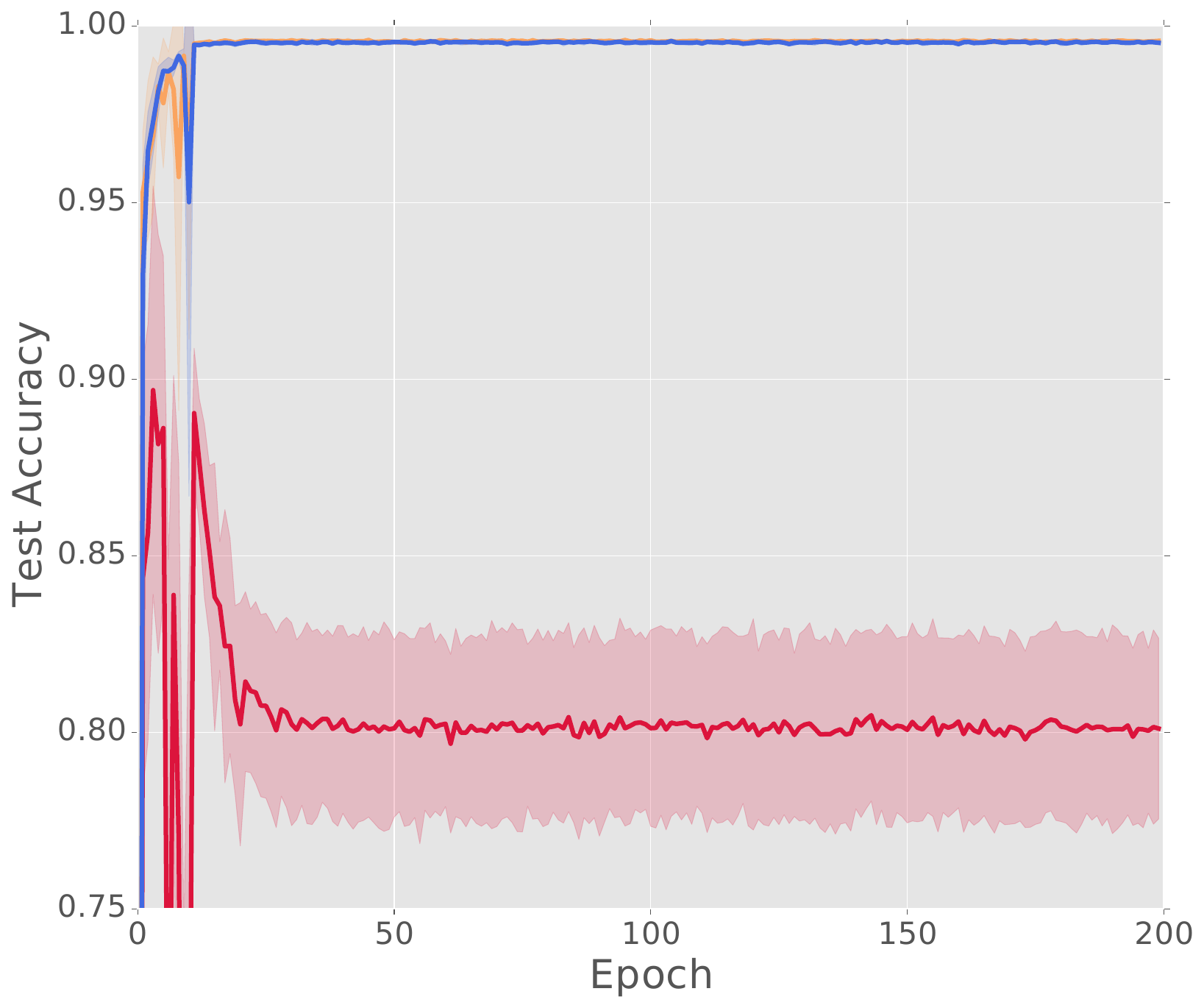}%
    \end{minipage}\\
    \begin{minipage}[c]{0.05\columnwidth}\centering\small \rotatebox[origin=c]{270}{|| CIFAR-10 ||} \end{minipage}%
    \begin{minipage}[c]{0.9\textwidth}
        \includegraphics[width=0.33\textwidth]{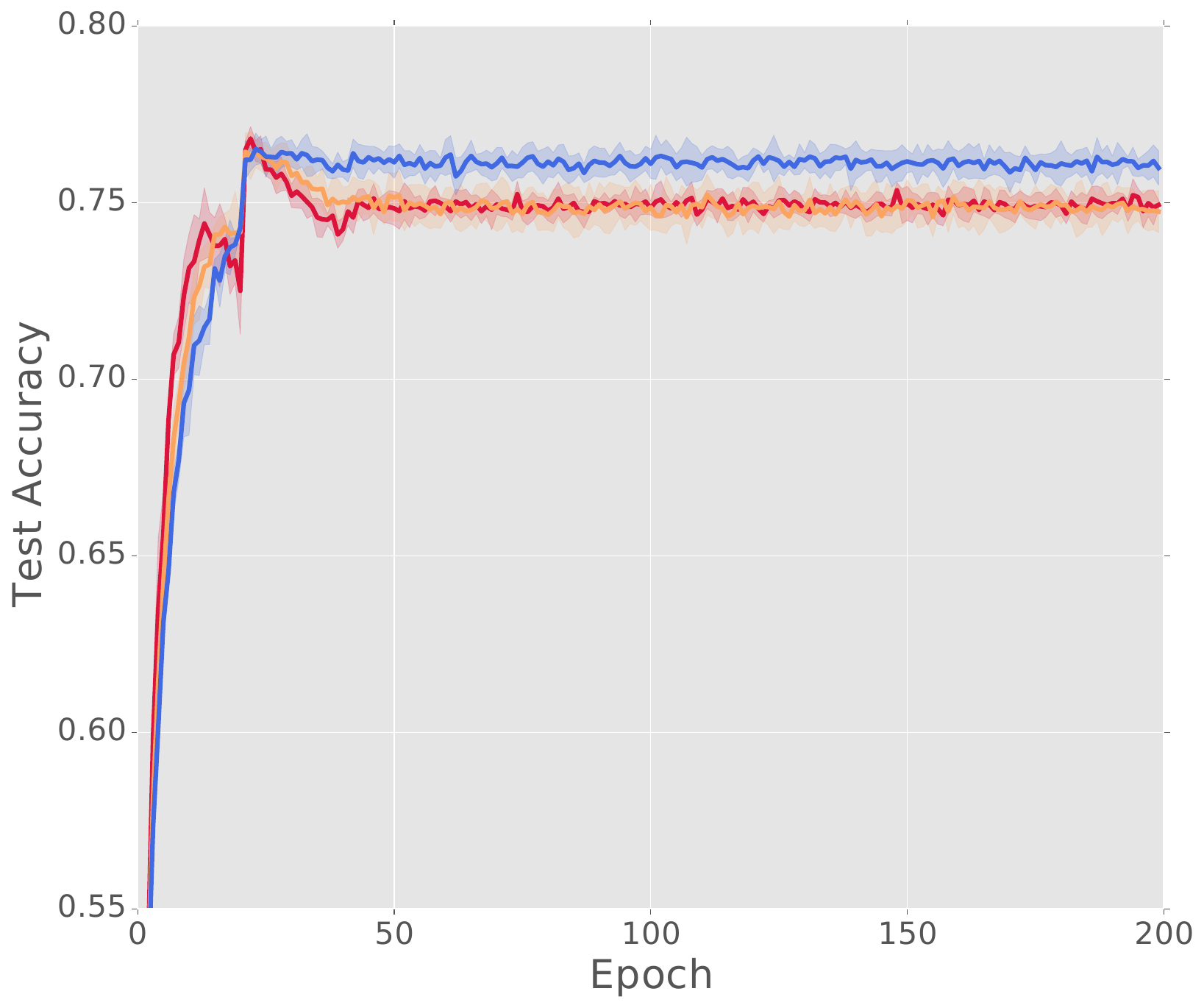}%
        \includegraphics[width=0.33\textwidth]{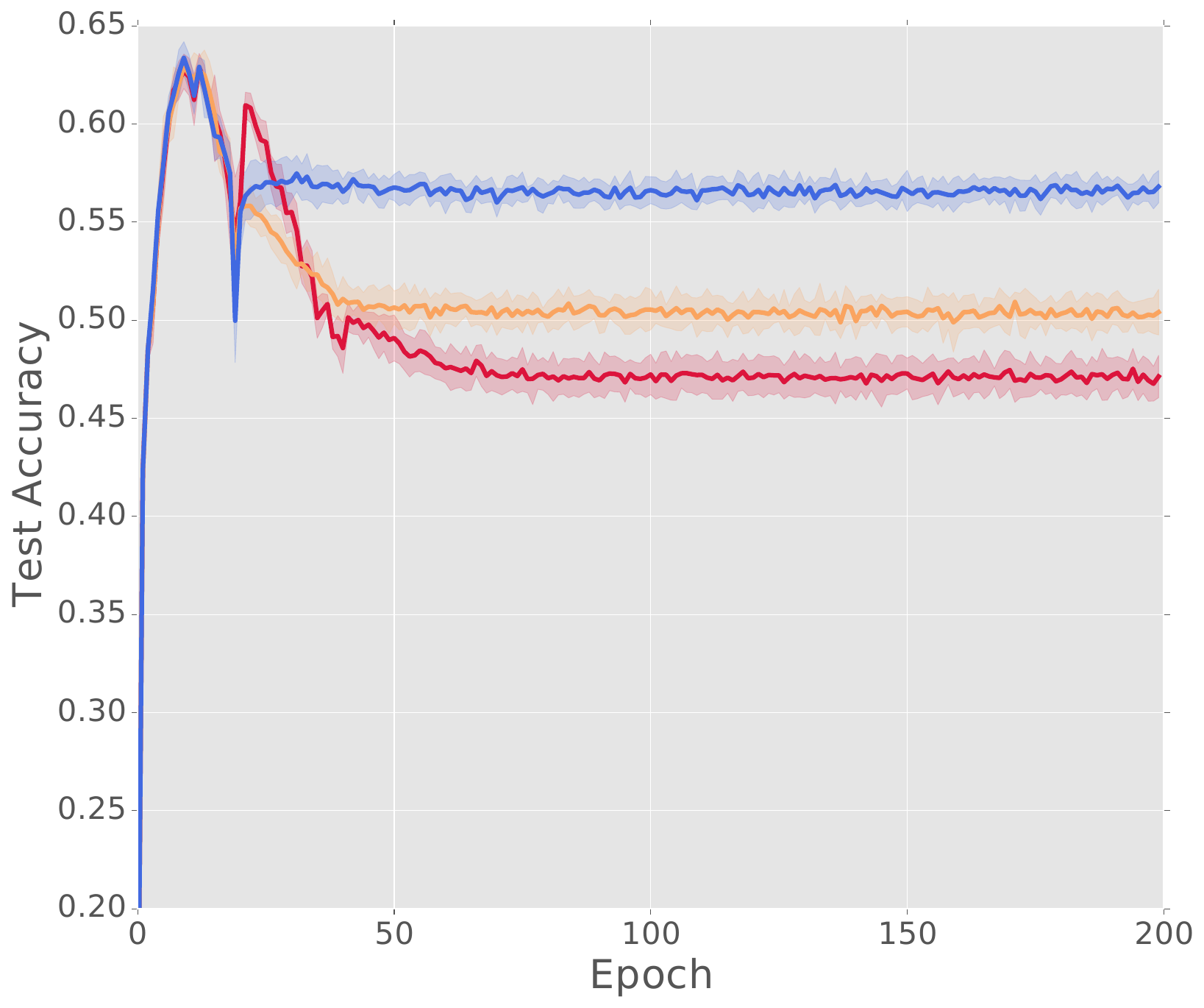}%
        \includegraphics[width=0.33\textwidth]{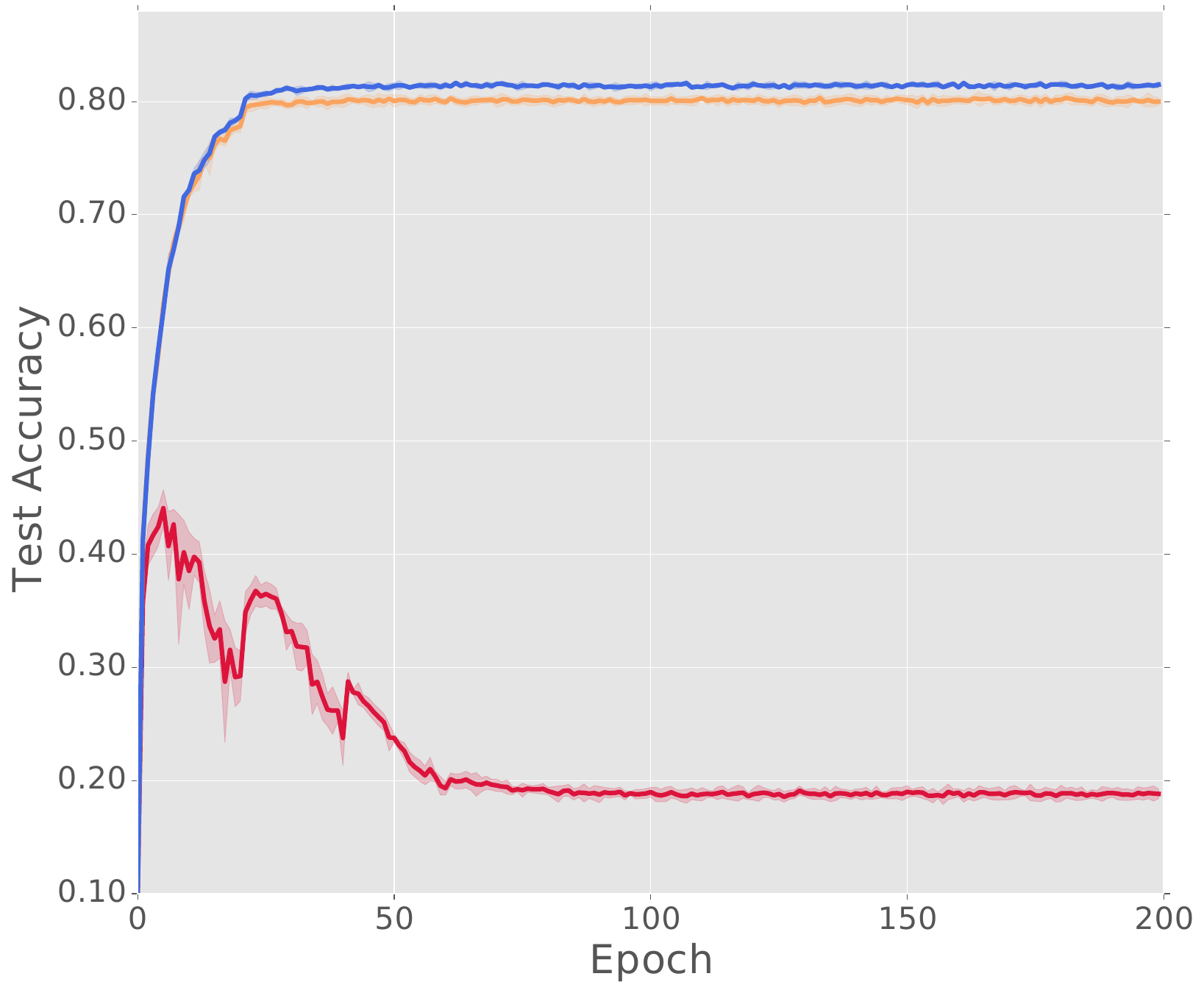}%
    \end{minipage}\\
    \vspace{-1ex}%
    \caption{Accuracy curves of training deep networks using the three learning methods in SET2.}
    \label{fig:back-corr-mnist-cifar10}
    \vspace{-1ex}%
\end{figure*}

We verify the effectiveness of SIGUA$\SL$ and SIGUA$\BC$ on noisy MNIST, CIFAR-10, CIFAR-100 and NEWS following \citet{han2018coteaching}.
Three noises are considered:
\vspace{-1ex}%
\begin{itemize}
    \item under \emph{symmetry-20\%}, $[T]_{i,i}=0.8$ and $\forall j\neq i,[T]_{i,j}=0.2/9$, where the \emph{intact-vs-flipped margin} is 0.78;
    \item under \emph{symmetry-50\%}, $[T]_{i,i}=0.5$ and $\forall j\neq i,[T]_{i,j}=0.5/9$, where the intact-vs-flipped margin is 0.44;
    \item under \emph{pair-45\%}, $[T]_{i,i}=0.55$, $[T]_{i,i\bmod10+1}=0.45$, and other entries are 0, where the margin is 0.10.
\end{itemize}
\vspace{-1ex}%
Thus, the noises move from easy to harder until very hard.
Additionally, we test SIGUA$\SL$ and SIGUA$\BC$ on the more challenging \emph{open-set} setting \citep{wang2018iterative,lee2019robust} by replacing CIFAR-10 images with SVHN images while keeping the labels of those ``mislabeled'' data intact.

According to the learning methods being involved, the experiments can be divided into two sets.
\emph{SET1} involves
\vspace{-1ex}%
\begin{itemize}
    \item \emph{standard training} with $\Cg\equiv1,\Cb\equiv0$,
    \item \emph{self-teaching} with $\Cg$ as Eq.~\eqref{eq:cond-good-small-loss} and $\Cb\equiv0$,
    \item SIGUA$\SL$ with $\Cg$ as Eq.~\eqref{eq:cond-good-small-loss} and $\Cb$ as Eq.~\eqref{eq:cond-bad-small-loss}.
\end{itemize}
\vspace{-1ex}%
The first method is $\fB$ in this set where the surrogate loss $\ell$ is softmax cross-entropy loss.
For $\rho(t)$ in Eq.~\eqref{eq:rho}, $T_k=10$ and $\epsilon$ is given as its true value; $\delta(t)$ is a constant independent of $t$ and depends only on the dataset.
\emph{SET2} involves%
\vspace{-1ex}%
\begin{itemize}
    \item \emph{backward correction}~(BC) with $\Cg\equiv1,\Cb\equiv0$,
    \item \emph{non-negative backward correction}~(nnBC) with $\Cg$ as Eq.~\eqref{eq:cond-good-back-corr} and $\Cb\equiv0$,
    \item SIGUA$\BC$ with $\Cg$ as Eq.~\eqref{eq:cond-good-back-corr} and $\Cb$ as Eq.~\eqref{eq:cond-bad-back-corr}.
\end{itemize}
\vspace{-1ex}%
The first method is $\fB$ in this set where $\ell$ in Eq.~\eqref{eq:loss-back-corr} is again softmax cross-entropy loss.
$T$ is given as its true value for constructing $\ellb$ in Eq.~\eqref{eq:loss-back-corr}.
Note that our experiments are proof-of-concept, and the baselines are just chosen for this purpose.
In principle, SIGUA can robustify other methods \citep[e.g.,][]{reed2014training,goldberger2016training,hendrycks2019using} provided that we could distinguish desired and undesired memorization conceptually.

The six learning methods are implemented using PyTorch.
In SET1, $\fO$ is Adam \citep{kingma2014adam} in its default,%
\footnote{With learning rate \emph{lr} as 0.001 and coefficients for computing running averages of gradient and its square \emph{betas} as (0.9, 0.999).} and the number of epochs is 200 with batch size $\nmb$ as 128;
the learning rate is linearly decayed to 0 from epoch 80 to 200.
We set $\gamma=0.01$ for all cases, except that $\gamma=0.001$ for pair-45\% of MNIST.
SET2 is a bit complicated:
\vspace{-1ex}%
\begin{itemize}
    \item for MNIST, $\fO$ is Adam with betas as (0.9, 0.1), and lr is divided by 10 every 10 epochs;
    \item for CIFAR-10, $\fO$ is SGD with momentum as 0.9, and lr is divided by 10 every 20 epochs;
    \item other hyperparameters have the same values as in SET1.
\end{itemize}
\vspace{-1ex}%
We simply set $\gamma=1.0$ for all cases.
Neural network architectures for benchmark datasets are given in Appendix~\ref{sec:details}.
Data augmentation is excluded from consideration \citep[cf.][]{ma2018dimensionality,zhang2018generalized}, as our experiments are proof-of-concept.
Due to the limited space, the experiments on CIFAR-100 and NEWS are completely deferred to Appendix~\ref{sec:more-experiments}.
All the experiments are repeated five times and the mean accuracy with standard deviation is recorded for each method in SET1 or SET2.

\vspace{-2ex}%
\paragraph{SET1 results.}
Figure~\ref{fig:small-loss-mnist-cifar10} shows the accuracy curves of the three methods in SET1 (MNIST in the top and CIFAR-10 in the bottom).
We can clearly see in Figure~\ref{fig:small-loss-mnist-cifar10} that models learned patterns first, and hence a robust learning method should be able to stop (or alleviate) the accuracy decrease.
On this point, SIGUA$\SL$ stopped the decrease in Standard and Self-Teach under two symmetry cases, and alleviated the decrease under pair-45\%, on MNIST.%
\footnote{The model on MNIST is exactly same as CIFAR-10/100---a 9-layer CNN---which is more than needed. This makes the accuracy very high in the beginning, while the overfitting is owing to not only noisy labels but also excess expressive power.}
SIGUA$\SL$ did a particularly good job on CIFAR-10, where it successfully made the accuracy continue to increase without a remarkable decrease, which indicates that SIGUA$\SL$ is superior to early stopping.
Table~\ref{tab:openset} shows the average accuracy under the open-set noise and we can see SIGUA$\SL$ outperformed Self-Teach significantly.
In summary, SIGUA consistently improved Self-Teach under the easy, harder, and very hard noises (note that the scales of y-axis are different), and the improvements were always significant.

\begin{table}[t]
    \vspace{-2ex}%
    \centering\small
    \caption{Average test accuracy (in \%) over the last ten epochs on CIFAR-10 under 40\% open-set noise from SVHN.}
    \label{tab:openset}
    \begin{tabular*}{\columnwidth}{ccc|ccc}
        \toprule
        Standard & Self & SIGUA$\SL$ & BC & nnBC & SIGUA$\BC$ \\
        \midrule
        56.44 & 79.72 & \textbf{81.31} & 52.03 & 73.39 & 74.33 \\
        \bottomrule
    \end{tabular*}
    \vspace{-1ex}%
\end{table}

\vspace{-1ex}%
\paragraph{SET2 results.}
Figure~\ref{fig:back-corr-mnist-cifar10} shows the accuracy curves of the three methods in SET2.
Surprisingly, models trained with BC still learned patterns first, even though $\ell$ was corrected into $\ellb$.
This implies there should be some other cause of overfitting (since Eq.~\eqref{eq:guarantee-x-back-corr} holds for any $x$), and indeed the cause is negative $\ellb$ on certain $(x_i,\ytil_i)$.
In Figure~\ref{fig:back-corr-mnist-cifar10}, nnBC was sometimes good enough but sometimes not enough to stop or alleviate the accuracy decrease, since nnBC tries to ignore rather than fix any negative loss.
SIGUA$\BC$ stopped the decrease in 5 cases and alleviated it in 1 case by fixing negative losses.
In Table~\ref{tab:openset}, we can also see that SIGUA$\BC$ outperformed BC and nnBC significantly.
In order to sum up, SIGUA consistently improved BC/nnBC under different noises, and the improvements were often significant.%
\footnote{Notice that nnBC is also a method proposed in this paper.}

\vspace{-1ex}%
\paragraph{Comparing results from SET1 and SET2.}
Note that $\epsilon$ or $T$ is given to SET1 or SET2 methods, and thus there is no error in estimating the label corruption process.
We can roughly see on MNIST, the performance of SIGUA$\SL$ and SIGUA$\BC$ were very close under two symmetry cases, but SIGUA$\BC$ was superior under pair-45\%;
on CIFAR-10, the performance of SIGUA$\BC$ was inferior to SIGUA$\SL$ under symmetry cases and again superior under pair-45\%.%
\footnote{The comparison is slightly unfair, since SET1 methods only knew $\epsilon$ whereas SET2 methods fully knew $T$.}

This is because the small-loss criterion is more reliable under symmetry noises than pair noises, where the reliability is determined by the intact-vs-flipped margin more than by the noise level $\epsilon$.
On the other hand, Eq.~\eqref{eq:loss-back-corr} for constructing $\ellb$ is equally reliable under different noises as long as the noise belongs to CCN.
This explains why SET2 methods were remarkably outperformed by SET1 methods under open-set noise---this noise does not even belong to the label noise, let alone CCN.
Actually, sample selection is a bit more general than loss correction and label correction, in a sense that it serves as corrections for not only flipping $y$ in label noise but also replacing $x$ in open-set noise.

Last but not least, comparing SET1 and SET2, we can find that desired memorization is not a concept definitely associated with intact data, and then neither is undesired memorization definitely associated with flipped data.

\begin{table}[t]
    \vspace{-2ex}%
    \centering\small
    \caption{Average test accuracy w.~std dev (in \%) over the last ten epochs of supervised learning, SIGUA$\SL$ and SIGUA$\BC$.}
    \label{tab:perf-gap}
    \begin{tabular*}{\columnwidth}{l*{3}{@{\extracolsep{\fill}}c}}
        \toprule
        MNIST & Symmetry-20\% & Symmetry-50\% & Pair-45\% \\
        \midrule
        Supervised & 99.61 (0.02) & 99.61 (0.02) & 99.61 (0.02) \\
        SIGUA$\SL$ & 98.91 (0.19) & 98.10 (0.30) & 89.37 (0.82) \\
        SIGUA$\BC$ & 99.42 (0.10) & 97.73 (0.05) & 99.47 (0.02) \\
        \bottomrule
    \end{tabular*}
    \vspace{-2ex}%
    \caption{Average test accuracy w.~std dev (in \%) over the last ten epochs of SIGUA$\BC$ and robust-loss-based learning methods.}
    \label{tab:robust-loss}
    \begin{tabular*}{\columnwidth}{l*{3}{@{\extracolsep{\fill}}c}}
        \toprule
        MNIST & Symmetry-20\% & Symmetry-50\% & Pair-45\% \\
        \midrule
        SIGUA$\BC$ & 99.42 (0.10) & 97.73 (0.05) & 99.47 (0.02) \\
        Huber & 93.61 (0.25) & 65.38 (0.33) & 56.48 (0.67) \\
        Log-sum & 94.35 (0.12) & 67.46 (0.40) & 57.38 (0.33) \\
        \bottomrule
    \end{tabular*}
    \vspace{-1ex}%
\end{table}

\vspace{-1ex}%
\paragraph{Comparison with supervised learning.}
Next, we investigate the performance gap between SIGUA and the oracle \emph{supervised learning with clean labels}.
In supervised learning, we train the same model using the same optimizer but on label-noise-free training data, and thus its performance upper bounds the performance of any learning method, no matter existing or to be proposed in the future, for learning with noisy labels.
The results are shown in Table~\ref{tab:perf-gap}, where SIGUA approximately approached the performance of supervised learning and achieved a rather small performance gap.
It is not surprised, as MNIST is an easy dataset, CCN is a relatively easy noise to correct, and $T$ is given.
In this research area, instance-dependent noise is worth us to pay more attention; within CCN the bottleneck/focus is how to estimate $T$ more and more accurately \citep{xia2019anchor}.

\vspace{-1ex}%
\paragraph{Comparison with robust losses.}
In the end, $\ellb$ is compared with \emph{Huber loss} and \emph{log-sum loss} for \emph{robust regression} \cite{candes2008enhancing}.
To make use of them, let $\by$ and $\bytil$ be the one-hot vectors of $y$ and $\ytil$, and $\ell(f(x),\bytil)$ be the sum of Huber/log-sum losses from $k$ dimensions ($f$ itself is vector-valued).
The assumed noise model is \emph{additive} by these losses: $\by$ and $\bytil$ are continuous and their difference $\beps$ is sampled from $(1-\epsilon)\cN(\bzero,\sigma I)+\epsilon\cN(\bzero,\sigma'I)$ where $\cN$ denotes the multivariate normal distribution, $\bzero$ denotes the all-zero vector in $\bbR^k$, and $\sigma\ll\sigma'$ for covariance matrices.
The results are shown in Table~\ref{tab:robust-loss}, where two robust losses notably failed under symmetry-50\% and pair-45\%.
It is as expected, since they are specially designed to be robust to outliers or similar additive noises, whereas $\ellb$ is specially designed to be robust to flipping noises.

\section{Conclusions}

We presented in this paper a versatile approach to learning with noisy labels called SIGUA.
By carefully distinguishing desired and undesired memorization, SIGUA was successful in robustifying two typical base learning methods: self-teaching from sample selection, and backward correction from loss correction.
We demonstrated through experiments that two enhanced methods can result in significant improvements.
In general, SIGUA can be applied to other methods or even other problem settings like learning from \emph{similarity-unlabeled} data \citep[e.g.,][]{bao2018classification} and \emph{pairwise comparison} data \citep[e.g.,][]{xu2019uncoupled}.

SIGUA exhibits pulling optimization back for generalization in learning with noisy labels.
There should be a trade-off between them implemented as an equilibrium between gradient descent and ascent, and then the model will travel between (uncountably infinite) reasonably good solutions, where the goodness is in the sense of optimization instead of generalization.
See \citet{ishida2020flood} for a dedicated study of this phenomenon in supervised learning.

\section*{Acknowledgments}

BH was supported by the Early Career Scheme (ECS) through the Research Grants Council of Hong Kong under Grant No.22200720, HKBU Tier-1 Start-up Grant, HKBU CSD Start-up Grant and a RIKEN BAIHO Award.
IWT was supported by Australian Research Council under Grants DP180100106 and DP200101328.
MS was supported by the International Research Center for Neurointelligence (WPI-IRCN) at The University of Tokyo Institutes for Advanced Study.

\bibliography{sigua}

\begin{thebibliography}{59}
\providecommand{\natexlab}[1]{#1}
\providecommand{\url}[1]{\texttt{#1}}
\expandafter\ifx\csname urlstyle\endcsname\relax
  \providecommand{\doi}[1]{doi: #1}\else
  \providecommand{\doi}{doi: \begingroup \urlstyle{rm}\Url}\fi

\bibitem[Arpit et~al.(2017)Arpit, Jastrzebski, Ballas, Krueger, Bengio, Kanwal,
  Maharaj, Fischer, Courville, and Bengio]{arpit2017closer}
Arpit, D., Jastrzebski, S., Ballas, N., Krueger, D., Bengio, E., Kanwal, M.,
  Maharaj, T., Fischer, A., Courville, A., and Bengio, Y.
\newblock A closer look at memorization in deep networks.
\newblock In \emph{ICML}, 2017.

\bibitem[Bao et~al.(2018)Bao, Niu, and Sugiyama]{bao2018classification}
Bao, H., Niu, G., and Sugiyama, M.
\newblock Classification from pairwise similarity and unlabeled data.
\newblock In \emph{ICML}, 2018.

\bibitem[Berthon et~al.(2020)Berthon, Han, Niu, Liu, and
  Sugiyama]{berthon2020idn}
Berthon, A., Han, B., Niu, G., Liu, T., and Sugiyama, M.
\newblock Confidence scores make instance-dependent label-noise learning
  possible.
\newblock \emph{arXiv:2001.03772}, 2020.

\bibitem[Candes et~al.(2008)Candes, Wakin, and Boyd]{candes2008enhancing}
Candes, E., Wakin, M., and Boyd, S.
\newblock Enhancing sparsity by reweighted l1 minimization.
\newblock \emph{Journal of Fourier analysis and applications}, 14\penalty0
  (5-6):\penalty0 877--905, 2008.

\bibitem[Cheng et~al.(2020)Cheng, Liu, Ramamohanarao, and Tao]{cheng2020idn}
Cheng, J., Liu, T., Ramamohanarao, K., and Tao, D.
\newblock Learning with bounded instance- and label-dependent label noise.
\newblock In \emph{ICML}, 2020.

\bibitem[Chou et~al.(2020)Chou, Niu, Lin, and Sugiyama]{chou2020complementary}
Chou, Y.-T., Niu, G., Lin, H.-T., and Sugiyama, M.
\newblock Unbiased risk estimators can mislead: A case study of learning with
  complementary labels.
\newblock In \emph{ICML}, 2020.

\bibitem[{du Plessis} et~al.(2014){du Plessis}, Niu, and
  Sugiyama]{christo2014nips}
{du Plessis}, M.~C., Niu, G., and Sugiyama, M.
\newblock Analysis of learning from positive and unlabeled data.
\newblock In \emph{NeurIPS}, 2014.

\bibitem[Duchi et~al.(2011)Duchi, Hazan, and Singer]{duchi2011adagrad}
Duchi, J., Hazan, E., and Singer, Y.
\newblock Adaptive subgradient methods for online learning and stochastic
  optimization.
\newblock \emph{Journal of Machine Learning Research}, 12:\penalty0 2121--2159,
  2011.

\bibitem[Feng et~al.(2020)Feng, Kaneko, Han, Niu, An, and
  Sugiyama]{feng2020complementary}
Feng, L., Kaneko, T., Han, B., Niu, G., An, B., and Sugiyama, M.
\newblock Learning with multiple complementary labels.
\newblock In \emph{ICML}, 2020.

\bibitem[Glorot \& Bengio(2010)Glorot and Bengio]{glorot2010understanding}
Glorot, X. and Bengio, Y.
\newblock Understanding the difficulty of training deep feedforward neural
  networks.
\newblock In \emph{AISTATS}, 2010.

\bibitem[Goldberger \& Ben-Reuven(2017)Goldberger and
  Ben-Reuven]{goldberger2016training}
Goldberger, J. and Ben-Reuven, E.
\newblock Training deep neural-networks using a noise adaptation layer.
\newblock In \emph{ICLR}, 2017.

\bibitem[Goodfellow et~al.(2016)Goodfellow, Bengio, and
  Courville]{goodfellow2016deep}
Goodfellow, I., Bengio, Y., and Courville, A.
\newblock \emph{Deep learning}.
\newblock MIT Press, 2016.

\bibitem[Han et~al.(2018{\natexlab{a}})Han, Yao, Niu, Zhou, Tsang, Zhang, and
  Sugiyama]{han2018masking}
Han, B., Yao, J., Niu, G., Zhou, M., Tsang, I., Zhang, Y., and Sugiyama, M.
\newblock Masking: A new perspective of noisy supervision.
\newblock In \emph{NeurIPS}, 2018{\natexlab{a}}.

\bibitem[Han et~al.(2018{\natexlab{b}})Han, Yao, Yu, Niu, Xu, Hu, Tsang, and
  Sugiyama]{han2018coteaching}
Han, B., Yao, Q., Yu, X., Niu, G., Xu, M., Hu, W., Tsang, I., and Sugiyama, M.
\newblock Co-teaching: Robust training of deep neural networks with extremely
  noisy labels.
\newblock In \emph{NeurIPS}, 2018{\natexlab{b}}.

\bibitem[He et~al.(2015)He, Zhang, Ren, and Sun]{he2015prelu}
He, K., Zhang, X., Ren, S., and Sun, J.
\newblock Delving deep into rectifiers: Surpassing human-level performance on
  imagenet classification.
\newblock In \emph{CVPR}, 2015.

\bibitem[He et~al.(2016)He, Zhang, Ren, and Sun]{he2016deep}
He, K., Zhang, X., Ren, S., and Sun, J.
\newblock Deep residual learning for image recognition.
\newblock In \emph{CVPR}, 2016.

\bibitem[Hendrycks et~al.(2018)Hendrycks, Mazeika, Wilson, and
  Gimpel]{hendrycks2018using}
Hendrycks, D., Mazeika, M., Wilson, D., and Gimpel, K.
\newblock Using trusted data to train deep networks on labels corrupted by
  severe noise.
\newblock In \emph{NeurIPS}, 2018.

\bibitem[Hendrycks et~al.(2019)Hendrycks, Lee, and Mazeika]{hendrycks2019using}
Hendrycks, D., Lee, K., and Mazeika, M.
\newblock Using pre-training can improve model robustness and uncertainty.
\newblock In \emph{ICML}, 2019.

\bibitem[Ioffe \& Szegedy(2015)Ioffe and Szegedy]{ioffe2015batchnorm}
Ioffe, S. and Szegedy, C.
\newblock Batch normalization: Accelerating deep network training by reducing
  internal covariate shift.
\newblock In \emph{ICML}, 2015.

\bibitem[Ishida et~al.(2017)Ishida, Niu, Hu, and
  Sugiyama]{ishida2017complementary}
Ishida, T., Niu, G., Hu, W., and Sugiyama, M.
\newblock Learning from complementary labels.
\newblock In \emph{NeurIPS}, 2017.

\bibitem[Ishida et~al.(2019)Ishida, Niu, Menon, and
  Sugiyama]{ishida2019complementary}
Ishida, T., Niu, G., Menon, A.~K., and Sugiyama, M.
\newblock Complementary-label learning for arbitrary losses and models.
\newblock In \emph{ICML}, 2019.

\bibitem[Ishida et~al.(2020)Ishida, Yamane, Sakai, Niu, and
  Sugiyama]{ishida2020flood}
Ishida, T., Yamane, I., Sakai, T., Niu, G., and Sugiyama, M.
\newblock Do we need zero training loss after achieving zero training error?
\newblock In \emph{ICML}, 2020.

\bibitem[Jiang et~al.(2018)Jiang, Zhou, Leung, Li, and
  Fei-Fei]{jiang2017mentornet}
Jiang, L., Zhou, Z., Leung, T., Li, L., and Fei-Fei, L.
\newblock Mentornet: Learning data-driven curriculum for very deep neural
  networks on corrupted labels.
\newblock In \emph{ICML}, 2018.

\bibitem[Kingma \& Ba(2015)Kingma and Ba]{kingma2014adam}
Kingma, D. and Ba, J.
\newblock Adam: A method for stochastic optimization.
\newblock In \emph{ICLR}, 2015.

\bibitem[Kiryo et~al.(2017)Kiryo, Niu, du~Plessis, and
  Sugiyama]{kiryo2017positive}
Kiryo, R., Niu, G., du~Plessis, M.~C., and Sugiyama, M.
\newblock Positive-unlabeled learning with non-negative risk estimator.
\newblock In \emph{NeurIPS}, 2017.

\bibitem[Krogh \& Hertz(1991)Krogh and Hertz]{krogh1991weightdecay}
Krogh, A. and Hertz, J.~A.
\newblock A simple weight decay can improve generalization.
\newblock In \emph{NeurIPS}, 1991.

\bibitem[Laine \& Aila(2017)Laine and Aila]{laine2017temporal}
Laine, S. and Aila, T.
\newblock Temporal ensembling for semi-supervised learning.
\newblock In \emph{ICLR}, 2017.

\bibitem[Lee et~al.(2019)Lee, Yun, Lee, Lee, Li, and Shin]{lee2019robust}
Lee, K., Yun, S., Lee, K., Lee, H., Li, B., and Shin, J.
\newblock Robust inference via generative classifiers for handling noisy
  labels.
\newblock In \emph{ICML}, 2019.

\bibitem[Liu \& Tao(2016)Liu and Tao]{liu2016classification}
Liu, T. and Tao, D.
\newblock Classification with noisy labels by importance reweighting.
\newblock \emph{IEEE Transactions on Pattern Analysis and Machine
  Intelligence}, 38\penalty0 (3):\penalty0 447--461, 2016.

\bibitem[Lu et~al.(2020)Lu, Zhang, Niu, and Sugiyama]{lu2019mitigating}
Lu, N., Zhang, T., Niu, G., and Sugiyama, M.
\newblock Mitigating overfitting in supervised classification from two
  unlabeled datasets: A consistent risk correction approach.
\newblock In \emph{AISTATS}, 2020.

\bibitem[Ma et~al.(2018)Ma, Wang, Houle, Zhou, Erfani, Xia, Wijewickrema, and
  Bailey]{ma2018dimensionality}
Ma, X., Wang, Y., Houle, M., Zhou, S., Erfani, S., Xia, S., Wijewickrema, S.,
  and Bailey, J.
\newblock Dimensionality-driven learning with noisy labels.
\newblock In \emph{ICML}, 2018.

\bibitem[Menon et~al.(2018)Menon, van Rooyen, and Natarajan]{menon2018idn}
Menon, A.~K., van Rooyen, B., and Natarajan, N.
\newblock Learning from binary labels with instance-dependent corruption.
\newblock \emph{Machine Learning}, 107:\penalty0 1561--1595, 2018.

\bibitem[Miyato et~al.(2019)Miyato, Maeda, Ishii, and
  Koyama]{miyato2019virtual}
Miyato, T., Maeda, S., Ishii, S., and Koyama, M.
\newblock Virtual adversarial training: a regularization method for supervised
  and semi-supervised learning.
\newblock \emph{IEEE Transactions on Pattern Analysis and Machine
  Intelligence}, 41\penalty0 (8):\penalty0 1979--1993, 2019.

\bibitem[Morgan \& Bourlard(1990)Morgan and Bourlard]{morgan1990earlystop}
Morgan, N. and Bourlard, H.
\newblock Generalization and parameter estimation in feedforward nets: Some
  experiments.
\newblock In \emph{NeurIPS}, 1990.

\bibitem[Nakkiran et~al.(2020)Nakkiran, Kaplun, Bansal, Yang, Barak, and
  Sutskever]{nakkiran2020ddd}
Nakkiran, P., Kaplun, G., Bansal, Y., Yang, T., Barak, B., and Sutskever, I.
\newblock Deep double descent: Where bigger models and more data hurt.
\newblock In \emph{ICLR}, 2020.

\bibitem[Natarajan et~al.(2013)Natarajan, Dhillon, Ravikumar, and
  Tewari]{natarajan2013learning}
Natarajan, N., Dhillon, I., Ravikumar, P., and Tewari, A.
\newblock Learning with noisy labels.
\newblock In \emph{NeurIPS}, 2013.

\bibitem[Niu et~al.(2016)Niu, {du Plessis}, Sakai, Ma, and
  Sugiyama]{niu2016nips}
Niu, G., {du Plessis}, M.~C., Sakai, T., Ma, Y., and Sugiyama, M.
\newblock Theoretical comparisons of positive-unlabeled learning against
  positive-negative learning.
\newblock In \emph{NeurIPS}, 2016.

\bibitem[Patrini et~al.(2017)Patrini, Rozza, Menon, Nock, and
  Qu]{patrini2017making}
Patrini, G., Rozza, A., Menon, A., Nock, R., and Qu, L.
\newblock Making deep neural networks robust to label noise: a loss correction
  approach.
\newblock In \emph{CVPR}, 2017.

\bibitem[Pennington et~al.(2014)Pennington, Socher, and
  Manning]{pennington2014glove}
Pennington, J., Socher, R., and Manning, C.~D.
\newblock {GloVe}: Global vectors for word representation.
\newblock In \emph{EMNLP}, 2014.

\bibitem[Reed et~al.(2015)Reed, Lee, Anguelov, Szegedy, Erhan, and
  Rabinovich]{reed2014training}
Reed, S., Lee, H., Anguelov, D., Szegedy, C., Erhan, D., and Rabinovich, A.
\newblock Training deep neural networks on noisy labels with bootstrapping.
\newblock In \emph{ICLR}, 2015.

\bibitem[Robbins \& Monro(1951)Robbins and Monro]{robbins51ams}
Robbins, H. and Monro, S.
\newblock A stochastic approximation method.
\newblock \emph{The Annals of Mathematical Statistics}, 22\penalty0
  (3):\penalty0 400--407, 1951.

\bibitem[Salimans et~al.(2016)Salimans, Goodfellow, Zaremba, Cheung, Radford,
  Chen, and Chen]{salimans2016improved}
Salimans, T., Goodfellow, I., Zaremba, W., Cheung, V., Radford, A., Chen, X.,
  and Chen, X.
\newblock Improved techniques for training gans.
\newblock In \emph{NeurIPS}, 2016.

\bibitem[Srivastava et~al.(2014)Srivastava, Hinton, Krizhevsky, Sutskever, and
  Salakhutdinov]{srivastava2014dropout}
Srivastava, N., Hinton, G., Krizhevsky, A., Sutskever, I., and Salakhutdinov,
  R.
\newblock Dropout: a simple way to prevent neural networks from overfitting.
\newblock \emph{Journal of Machine Learning Research}, 15\penalty0
  (56):\penalty0 1929--1958, 2014.

\bibitem[Suzuki(2019)]{suzuki2019adaptivity}
Suzuki, T.
\newblock Adaptivity of deep {ReLU} network for learning in {Besov} and mixed
  smooth {Besov} spaces: optimal rate and curse of dimensionality.
\newblock In \emph{ICLR}, 2019.

\bibitem[Tokui et~al.(2015)Tokui, Oono, Hido, and Clayton]{tokui2015chainer}
Tokui, S., Oono, K., Hido, S., and Clayton, J.
\newblock Chainer: a next-generation open source framework for deep learning.
\newblock In \emph{NeurIPS Workshop on Machine Learning Systems}, 2015.

\bibitem[Vapnik(1998)]{vapnik98SLT}
Vapnik, V.~N.
\newblock \emph{Statistical Learning Theory}.
\newblock John Wiley \& Sons, 1998.

\bibitem[Wang et~al.(2018)Wang, Liu, Ma, Bailey, Zha, Song, and
  Xia]{wang2018iterative}
Wang, Y., Liu, W., Ma, X., Bailey, J., Zha, H., Song, L., and Xia, S.
\newblock Iterative learning with open-set noisy labels.
\newblock In \emph{CVPR}, 2018.

\bibitem[Wang et~al.(2019)Wang, Ma, Chen, Luo, Yi, and
  Bailey]{wang2019symmetric}
Wang, Y., Ma, X., Chen, Z., Luo, Y., Yi, J., and Bailey, J.
\newblock Symmetric cross entropy for robust learning with noisy labels.
\newblock In \emph{ICCV}, 2019.

\bibitem[Welinder et~al.(2010)Welinder, Branson, Perona, and
  Belongie]{welinder2010multidimensional}
Welinder, P., Branson, S., Perona, P., and Belongie, S.
\newblock The multidimensional wisdom of crowds.
\newblock In \emph{NeurIPS}, 2010.

\bibitem[Xia et~al.(2019)Xia, Liu, Wang, Han, Gong, Niu, and
  Sugiyama]{xia2019anchor}
Xia, X., Liu, T., Wang, N., Han, B., Gong, C., Niu, G., and Sugiyama, M.
\newblock Are anchor points really indispensable in label-noise learning?
\newblock In \emph{NeurIPS}, 2019.

\bibitem[Xia et~al.(2020)Xia, Liu, Han, Wang, Gong, Liu, Niu, Tao, and
  Sugiyama]{xia2020parts}
Xia, X., Liu, T., Han, B., Wang, N., Gong, M., Liu, H., Niu, G., Tao, D., and
  Sugiyama, M.
\newblock Parts-dependent label noise: Towards instance-dependent label noise.
\newblock \emph{arXiv:2006.07836}, 2020.

\bibitem[Xiao et~al.(2015)Xiao, Xia, Yang, Huang, and Wang]{xiao2015learning}
Xiao, T., Xia, T., Yang, Y., Huang, C., and Wang, X.
\newblock Learning from massive noisy labeled data for image classification.
\newblock In \emph{CVPR}, 2015.

\bibitem[Xu et~al.(2015)Xu, Wang, Chen, and Li]{xu2015lrelu}
Xu, B., Wang, N., Chen, T., and Li, M.
\newblock Empirical evaluation of rectified activations in convolutional
  network.
\newblock In \emph{ICML Deep Learning Workshop}, 2015.

\bibitem[Xu et~al.(2019)Xu, Honda, Niu, and Sugiyama]{xu2019uncoupled}
Xu, L., Honda, J., Niu, G., and Sugiyama, M.
\newblock Uncoupled regression from pairwise comparison data.
\newblock In \emph{NeurIPS}, 2019.

\bibitem[Yao et~al.(2020)Yao, Liu, Han, Gong, Deng, Niu, and
  Sugiyama]{yao2020dual}
Yao, Y., Liu, T., Han, B., Gong, M., Deng, J., Niu, G., and Sugiyama, M.
\newblock Dual {T}: Reducing estimation error for transition matrix in
  label-noise learning.
\newblock \emph{arXiv:2006.07805}, 2020.

\bibitem[Yu et~al.(2018)Yu, Liu, Gong, and Tao.]{yu2018complementary}
Yu, X., Liu, T., Gong, M., and Tao., D.
\newblock Learning with biased complementary labels.
\newblock In \emph{ECCV}, 2018.

\bibitem[Yu et~al.(2019)Yu, Han, Yao, Niu, Tsang, and Sugiyama]{yu2019coplus}
Yu, X., Han, B., Yao, J., Niu, G., Tsang, I.~W., and Sugiyama, M.
\newblock How does disagreement help generalization against label corruption?
\newblock In \emph{ICML}, 2019.

\bibitem[Zhang et~al.(2017)Zhang, Bengio, Hardt, Recht, and
  Vinyals]{zhang2016understanding}
Zhang, C., Bengio, S., Hardt, M., Recht, B., and Vinyals, O.
\newblock Understanding deep learning requires rethinking generalization.
\newblock In \emph{ICLR}, 2017.

\bibitem[Zhang \& Sabuncu(2018)Zhang and Sabuncu]{zhang2018generalized}
Zhang, Z. and Sabuncu, M.
\newblock Generalized cross entropy loss for training deep neural networks with
  noisy labels.
\newblock In \emph{NeurIPS}, 2018.

\end{thebibliography}
\bibliographystyle{icml2020}

\clearpage
\onecolumn
\appendix

\section{Neural Network Architectures for Benchmark Datasets}
\label{sec:details}

\begin{table}[t]
    \centering
    \begin{minipage}[t]{0.34\textwidth}
        \centering\small
        \caption{CNN on MNIST and CIFAR-10/100.}
        \label{tab:models-cnn}
        \begin{tabular*}{\textwidth}{l@{\extracolsep{\fill}}c}
            \toprule
            \multirow{2}*{Input}
            & 28$\times$28 Gray Image \\
            & 32$\times$32 Color Image \\
            \midrule
            \multirow{5}*{Block 1}
            & Conv(3$\times$3, 128)-BN-LReLU \\
            & Conv(3$\times$3, 128)-BN-LReLU \\
            & Conv(3$\times$3, 128)-BN-LReLU \\
            & MaxPool(2$\times$2, stride = 2) \\
            & Dropout(p = 0.25) \\
            \midrule
            \multirow{5}*{Block 2}
            & Conv(3$\times$3, 256)-BN-LReLU \\
            & Conv(3$\times$3, 256)-BN-LReLU \\
            & Conv(3$\times$3, 256)-BN-LReLU \\
            & MaxPool(2$\times$2, stride = 2) \\
            & Dropout(p = 0.25) \\
            \midrule
            \multirow{4}*{Block 3}
            & Conv(3$\times$3, 512)-BN-LReLU \\
            & Conv(3$\times$3, 256)-BN-LReLU \\
            & Conv(3$\times$3, 128)-BN-LReLU \\
            & GlobalAvgPool(128) \\
            \midrule
            Score & Linear(128, 10 or 100) \\
            \bottomrule
        \end{tabular*}
    \end{minipage}\hfill
    \begin{minipage}[t]{0.32\textwidth}
        \centering\small
        \caption{CNN for open-set noise.}
        \label{tab:models-openset-cnn}
        \begin{tabular*}{\textwidth}{lc}
            \toprule
            Input & 32$\times$32 Color Image \\
            \midrule
            \multirow{3}*{Block 1}
            & Conv(3$\times$3, 64)-BN-LReLU \\
            & Conv(3$\times$3, 64)-BN-LReLU \\
            & MaxPool(2$\times$2) \\
            \midrule
            \multirow{3}*{Block 2}
            & Conv(3$\times$3, 128)-BN-LReLU \\
            & Conv(3$\times$3, 128)-BN-LReLU \\
            & MaxPool(2$\times$2) \\
            \midrule
            \multirow{3}*{Block 3}
            & Conv(3$\times$3, 196)-BN-LReLU \\
            & Conv(3$\times$3, 196)-BN-LReLU \\
            & MaxPool(2$\times$2) \\
            \midrule
            Score & Linear(256, 10) \\
            \bottomrule
        \end{tabular*}
    \end{minipage}\hfill
    \begin{minipage}[t]{0.3\textwidth}
        \centering\small
        \caption{1D CNN on NEWS.}
        \label{tab:models-news-cnn}
        \begin{tabular*}{\textwidth}{l@{\extracolsep{\fill}}c}
            \toprule
            Input & Sequence of Tokens \\
            \midrule
            Embed & 300D GloVe \\
            \midrule
            \multirow{2}*{Block 1}
            & Conv(3, 100)-ReLU \\
            & GlobalMaxPool(100) \\
            \midrule
            Score & Linear(100, 7) \\
            \bottomrule
        \end{tabular*}
        \vspace{30pt}%
        \caption{MLP on NEWS.}
        \label{tab:models-news-mlp}
        \begin{tabular*}{\textwidth}{lc}
            \toprule
            Input & Sequence of Tokens \\
            \midrule
            Embed & 300D GloVe \\
            \midrule
            \multirow{2}*{Block 1}
            & Linear(300, 300)-Softsign \\
            & Linear(300, 300)-Softsign \\
            \midrule
            Score & Linear(300, 2) \\
            \bottomrule
        \end{tabular*}
    \end{minipage}
    \vspace{-1ex}%
\end{table}

Table~\ref{tab:models-cnn} describes the 9-layer CNN \citep{laine2017temporal,miyato2019virtual} used on MNIST and CIFAR-10/100.
In fact, it has 9 convolutional layers but 19 trainable layers.
Table~\ref{tab:models-openset-cnn} describes the CNN used on CIFAR-10 under open-set noise.
It has 6 convolutional layers but 13 trainable layers.
Furthermore, the 1D CNN on NEWS for SET1 methods and MLP on NEWS for SET2 methods are given in Tables~\ref{tab:models-news-cnn} and \ref{tab:models-news-mlp} respectively.
Here, BN stands for \emph{batch normalization} layers \citep{ioffe2015batchnorm};
LReLU stands for \emph{Leaky ReLU} \citep{xu2015lrelu}, a special case of \emph{Parametric ReLU} \citep{he2015prelu};
GloVe stands for \emph{global vectors} for word representation \citep{pennington2014glove};
Softsign is an activation function which looks very similar to Tanh \citep{glorot2010understanding}.

Note that the 9-layer CNN is a standard and common practice in weakly supervised learning, including but not limited to semi-supervised learning \citep[e.g.,][]{laine2017temporal,miyato2019virtual} and noisy-label learning \citep[e.g.,][]{han2018coteaching}.
Actually, this CNN was not born in those areas---it came from \citet{salimans2016improved} where it served as the discriminator of GANs on CIFAR-10.
We decided to use this CNN, because then the experimental results are directly comparable with previous papers in the same area, and it would be crystal clear where the proposed methods stand in the area.

That being said, SIGUA can definitely achieve better performance if given better models.
In order to demonstrate this, let us take ResNet-18 \citep{he2016deep}, the smallest ResNet in \emph{torch vision model zoo} for example.
The experimental results are shown in Table~\ref{tab:resnet18}.
For each noise, we selected the better one among SIGUA$\SL$ and SIGUA$\BC$, and replaced the model with ResNet-18.
We can see from Table~\ref{tab:resnet18} that the improvements were very great by training bigger ResNet-18.

\begin{table}[h]
    \vspace{-2ex}%
    \centering\small
    \caption{Average test accuracy (in \%) over the last ten epochs on CIFAR-10.}
    \label{tab:resnet18}
    \begin{tabular*}{0.7\textwidth}{l*{3}{@{\extracolsep{\fill}}c}}
        \toprule
        & SIGUA$\SL$ under & SIGUA$\SL$ under & SIGUA$\BC$ \\
        & symmetry-20\% & symmetry-50\% & under pair-45\% \\
        \midrule
        9-layer CNN & 84.05 & 77.12 & 81.82 \\
        ResNet-18 & 89.41 & 81.96 & 89.56 \\
        Absolute Acc Increase & 5.36 & 4.84 & 7.74 \\
        Relative Err Reduction & 33.61 & 21.15 & 42.57 \\
        \bottomrule
    \end{tabular*}
    \vspace{-1ex}%
\end{table}

\section{More Experiments}
\label{sec:more-experiments}

\begin{figure*}[!h]
    \centering
    \begin{minipage}[c]{0.025\textwidth}~\end{minipage}%
    \begin{minipage}[c]{0.3\textwidth}\centering\small || Symmetry-20\% || \end{minipage}%
    \begin{minipage}[c]{0.3\textwidth}\centering\small || Symmetry-50\% || \end{minipage}%
    \begin{minipage}[c]{0.3\textwidth}\centering\small || Pair-45\% || \end{minipage}\\
    \begin{minipage}[c]{0.025\textwidth}\centering\small \rotatebox[origin=c]{270}{|| CIFAR-100 ||} \end{minipage}%
    \begin{minipage}[c]{0.9\textwidth}
        \includegraphics[width=0.33\textwidth]{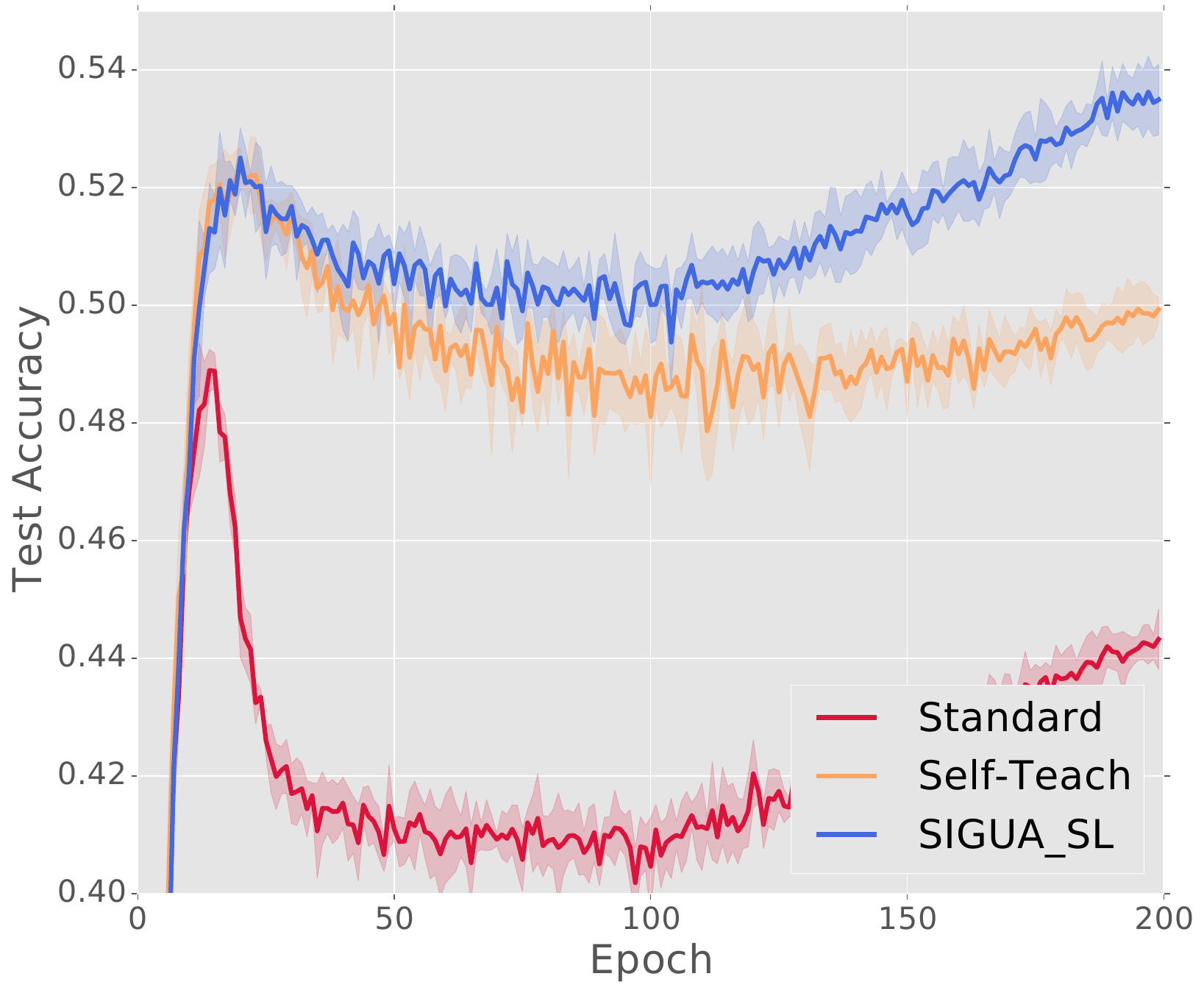}%
        \includegraphics[width=0.33\textwidth]{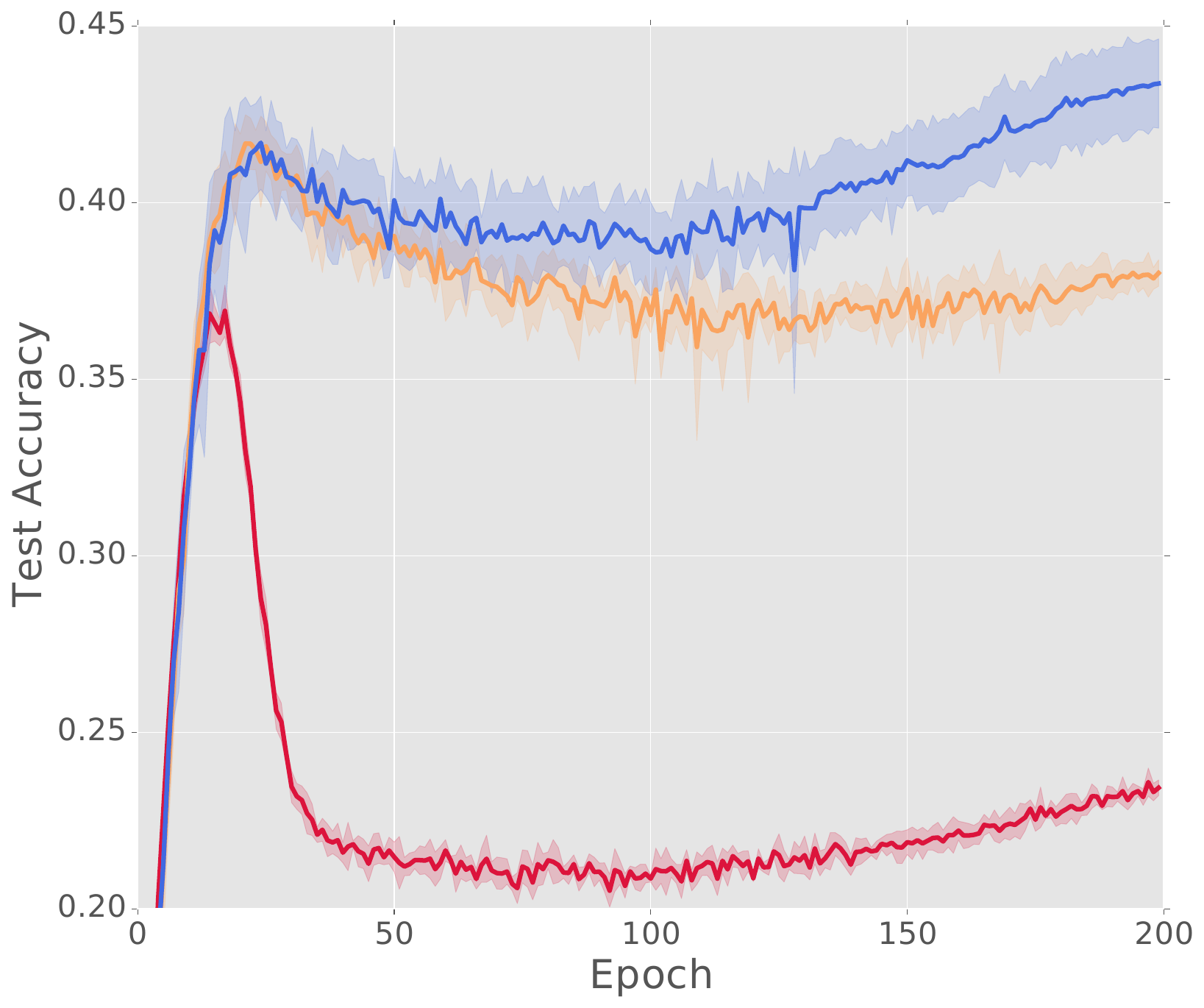}%
        \includegraphics[width=0.33\textwidth]{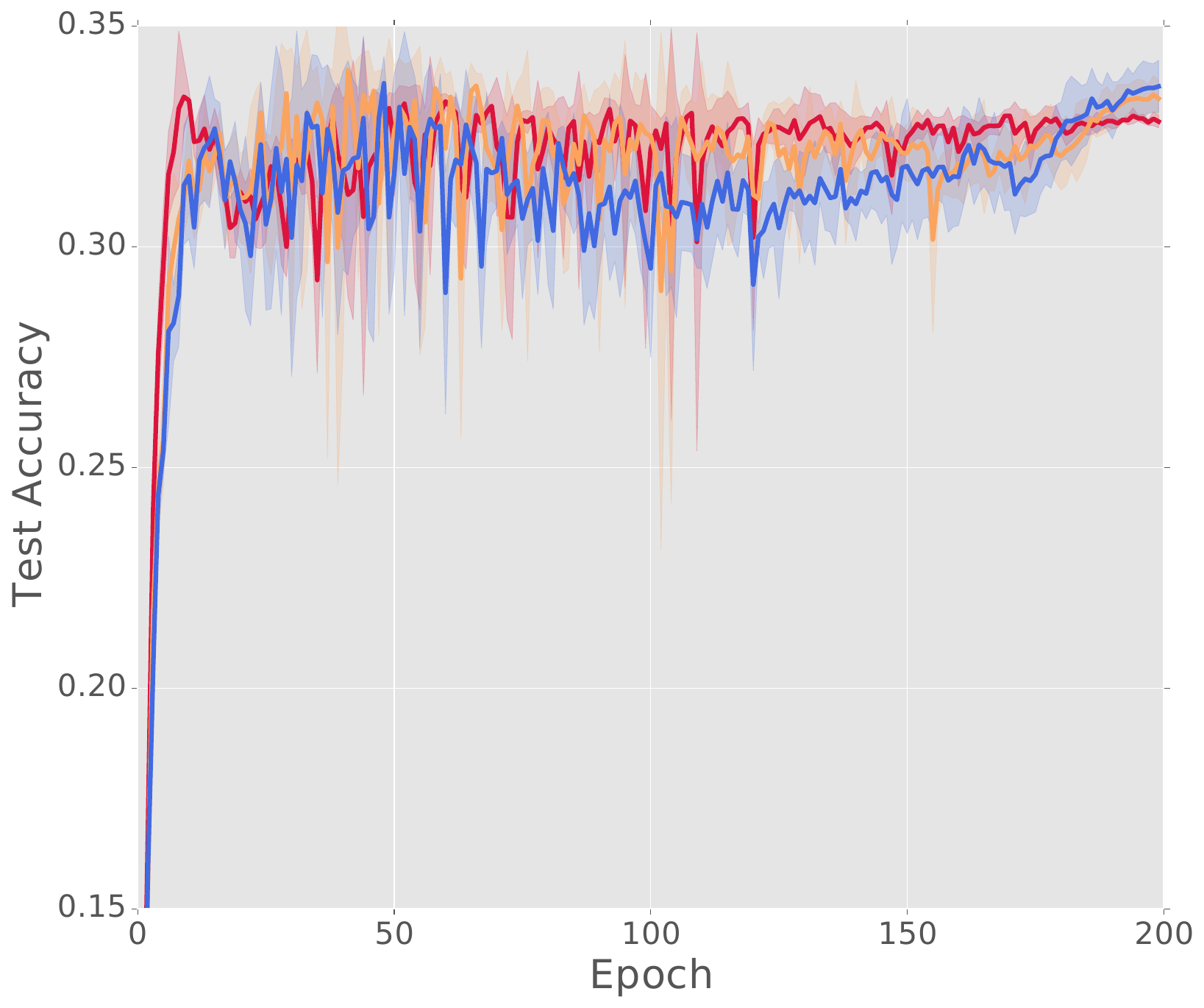}%
    \end{minipage}\\
    \begin{minipage}[c]{0.025\textwidth}\centering\small \rotatebox[origin=c]{270}{|| NEWS ||} \end{minipage}%
    \begin{minipage}[c]{0.9\textwidth}
        \includegraphics[width=0.33\textwidth]{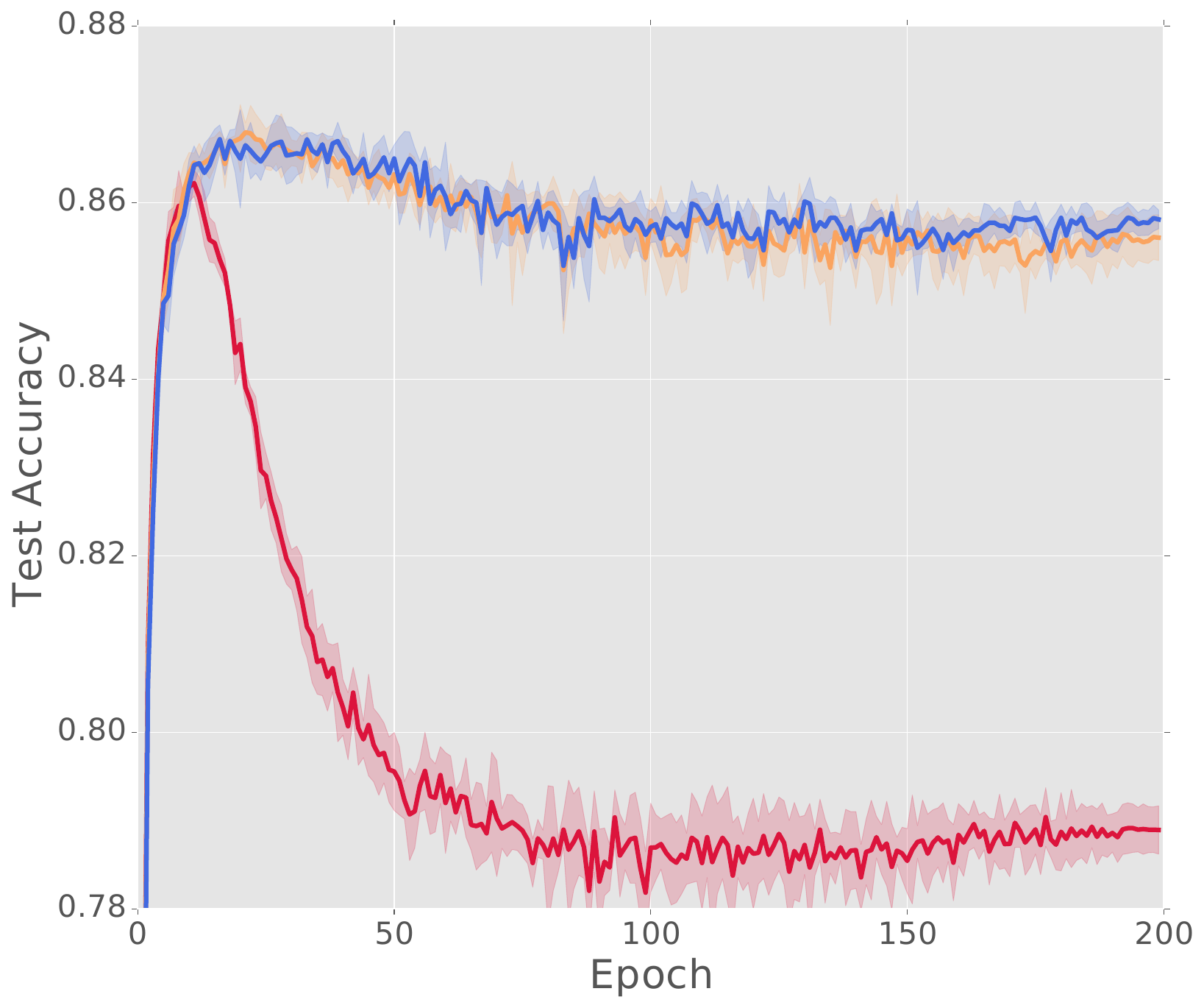}%
        \includegraphics[width=0.33\textwidth]{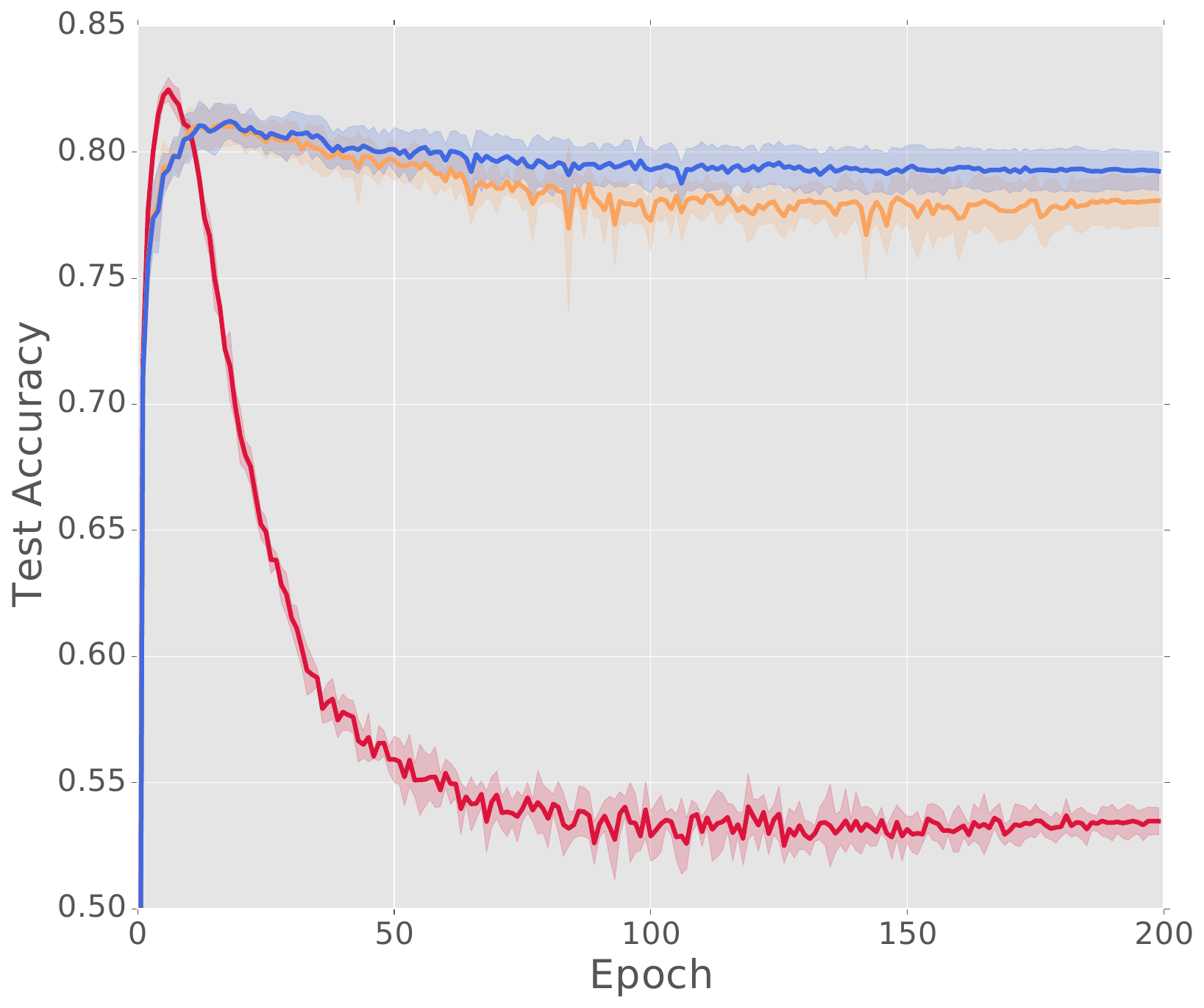}%
        \includegraphics[width=0.33\textwidth]{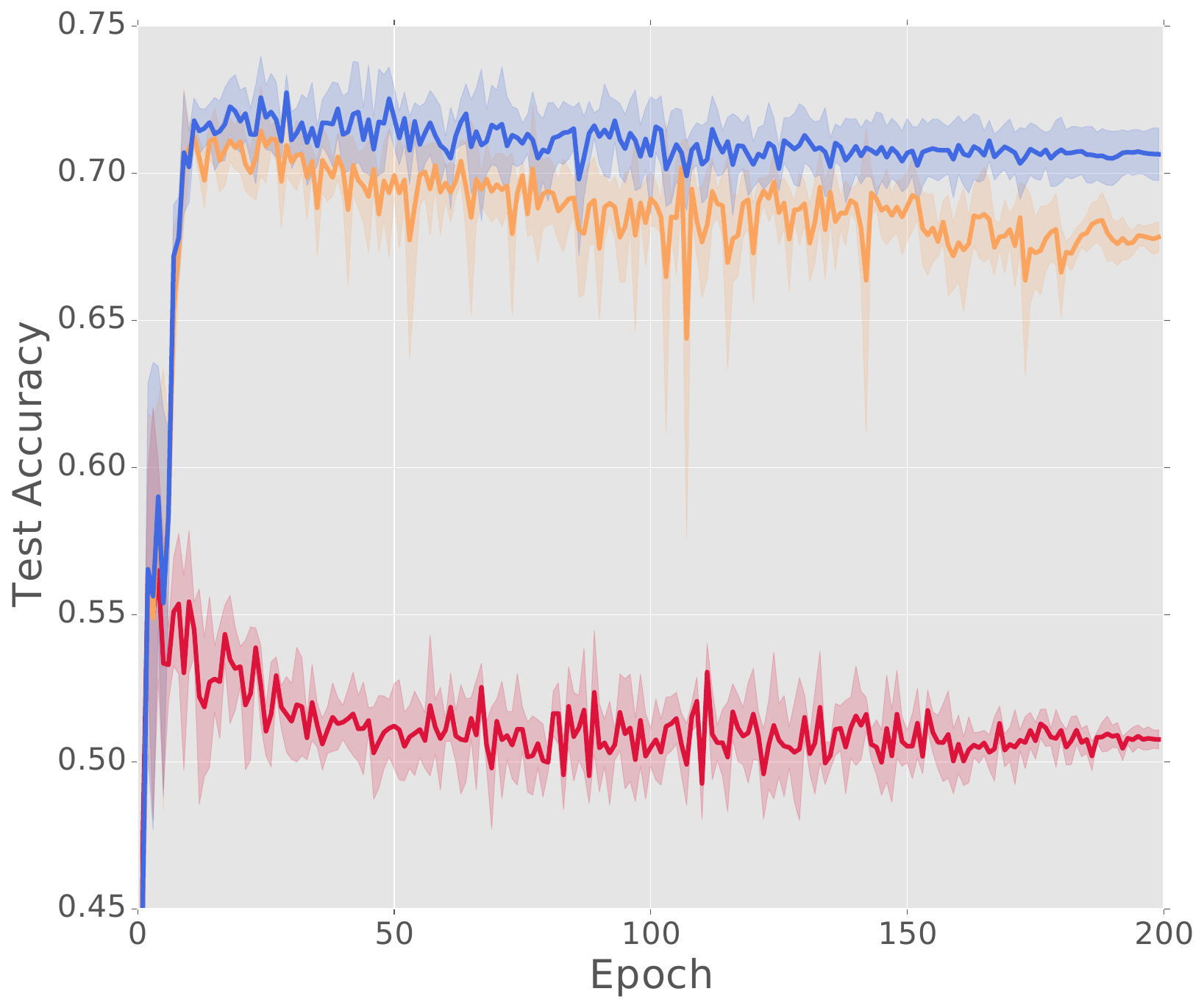}%
    \end{minipage}\\
    \vspace{-1ex}%
    \caption{Accuracy curves of training deep networks using the three learning methods in SET1.}
    \label{fig:small-loss-cifar100-news}
    \vspace{1ex}%
    \begin{minipage}[c]{0.025\textwidth}~\end{minipage}%
    \begin{minipage}[c]{0.3\textwidth}\centering\small || Symmetry-20\% || \end{minipage}%
    \begin{minipage}[c]{0.3\textwidth}\centering\small || Symmetry-50\% || \end{minipage}%
    \begin{minipage}[c]{0.3\textwidth}\centering\small || Pair-45\% || \end{minipage}\\
    \begin{minipage}[c]{0.025\textwidth}\centering\small \rotatebox[origin=c]{270}{|| CIFAR-100 ||} \end{minipage}%
    \begin{minipage}[c]{0.9\textwidth}
        \includegraphics[width=0.33\textwidth]{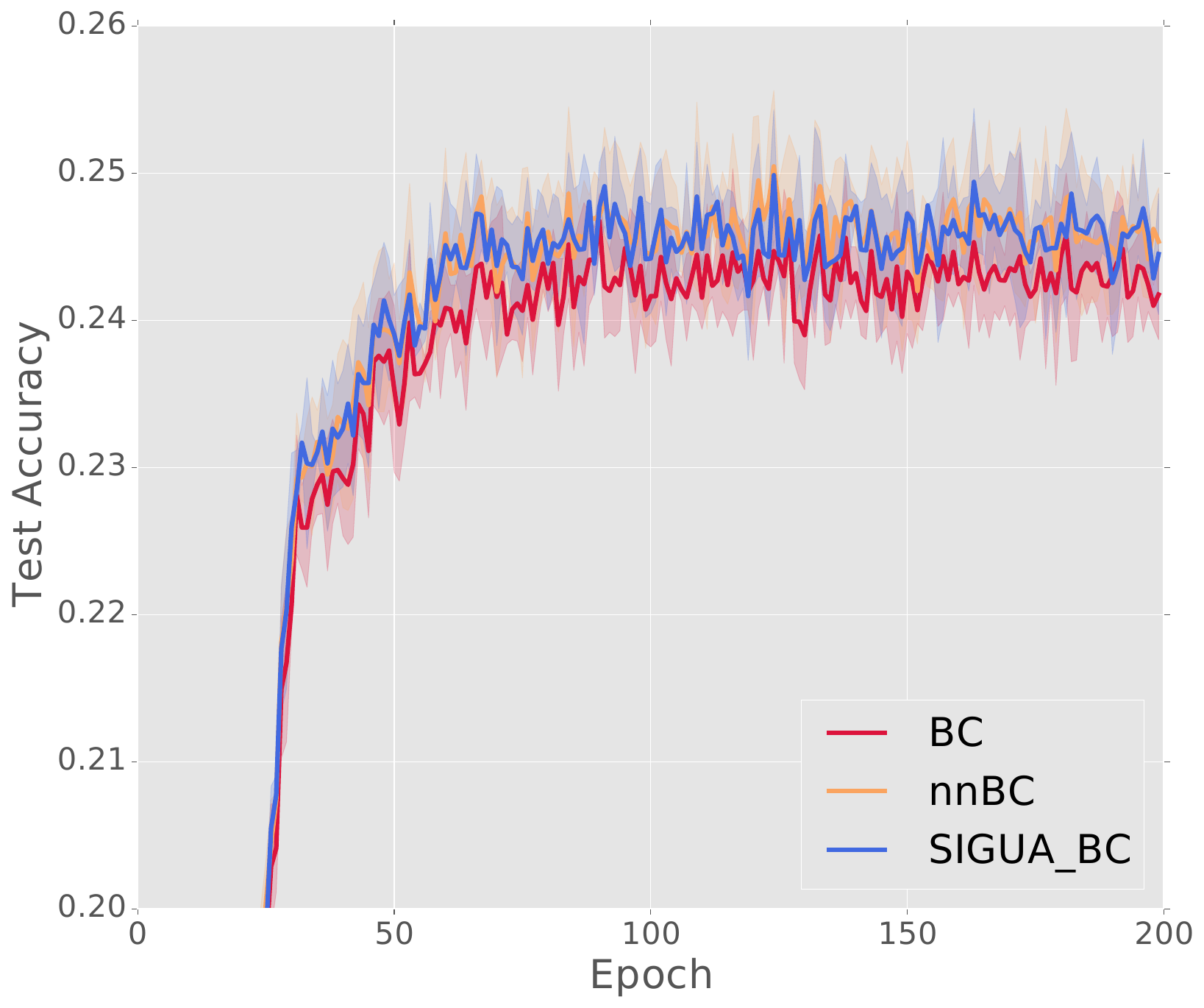}%
        \includegraphics[width=0.33\textwidth]{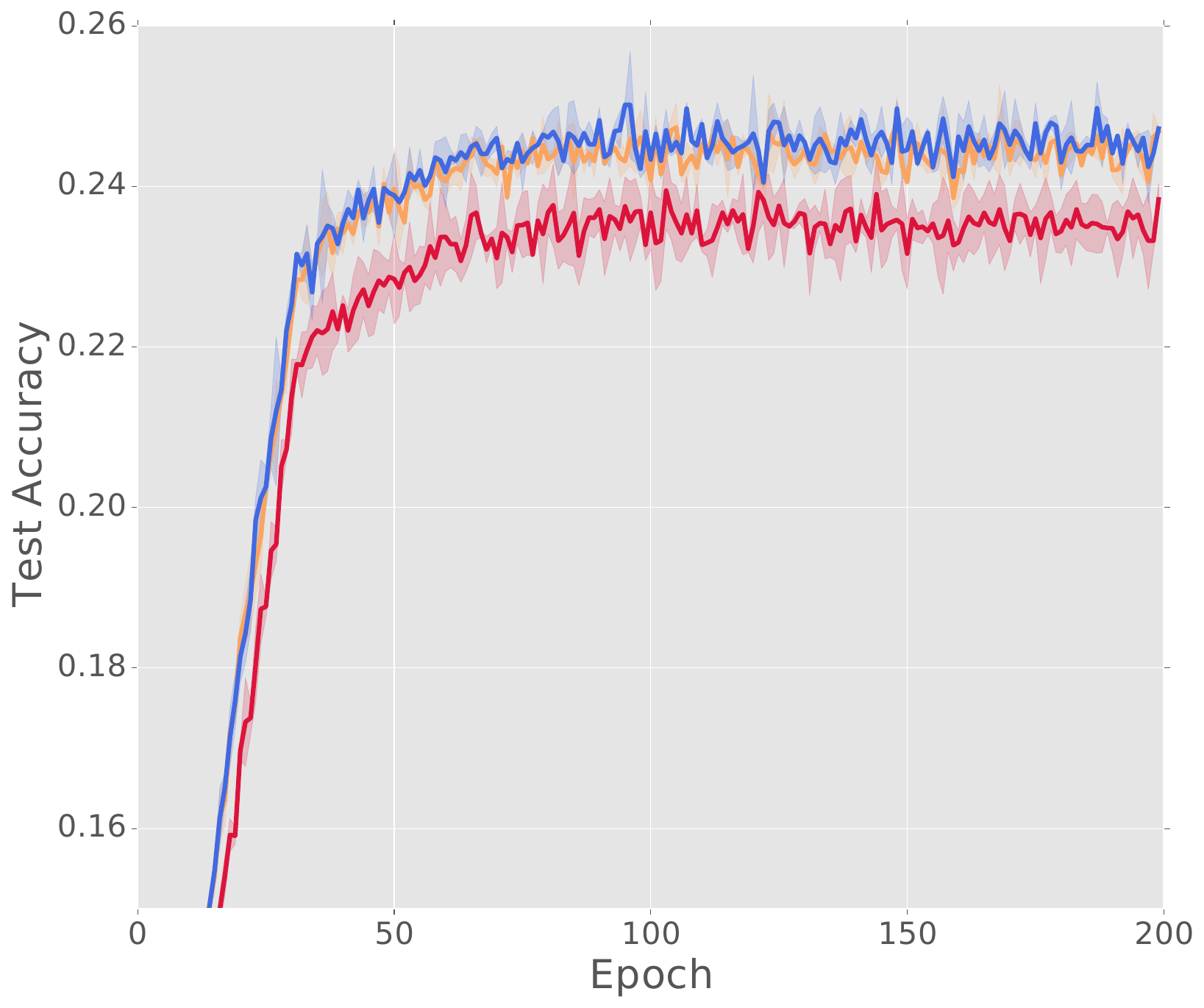}%
        \includegraphics[width=0.33\textwidth]{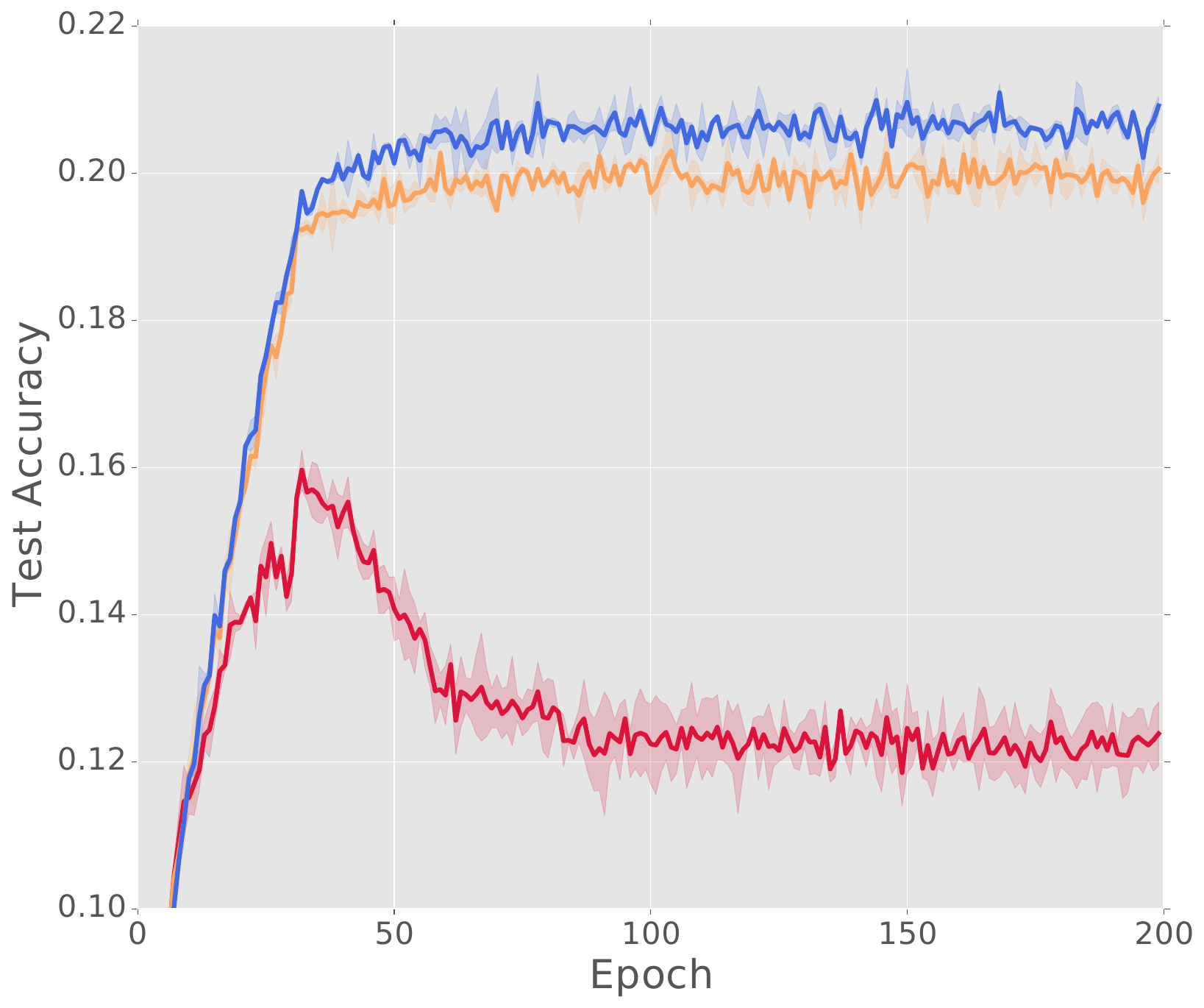}%
    \end{minipage}\\
    \begin{minipage}[c]{0.025\textwidth}\centering\small \rotatebox[origin=c]{270}{|| NEWS ||} \end{minipage}%
    \begin{minipage}[c]{0.9\textwidth}
        \includegraphics[width=0.33\textwidth]{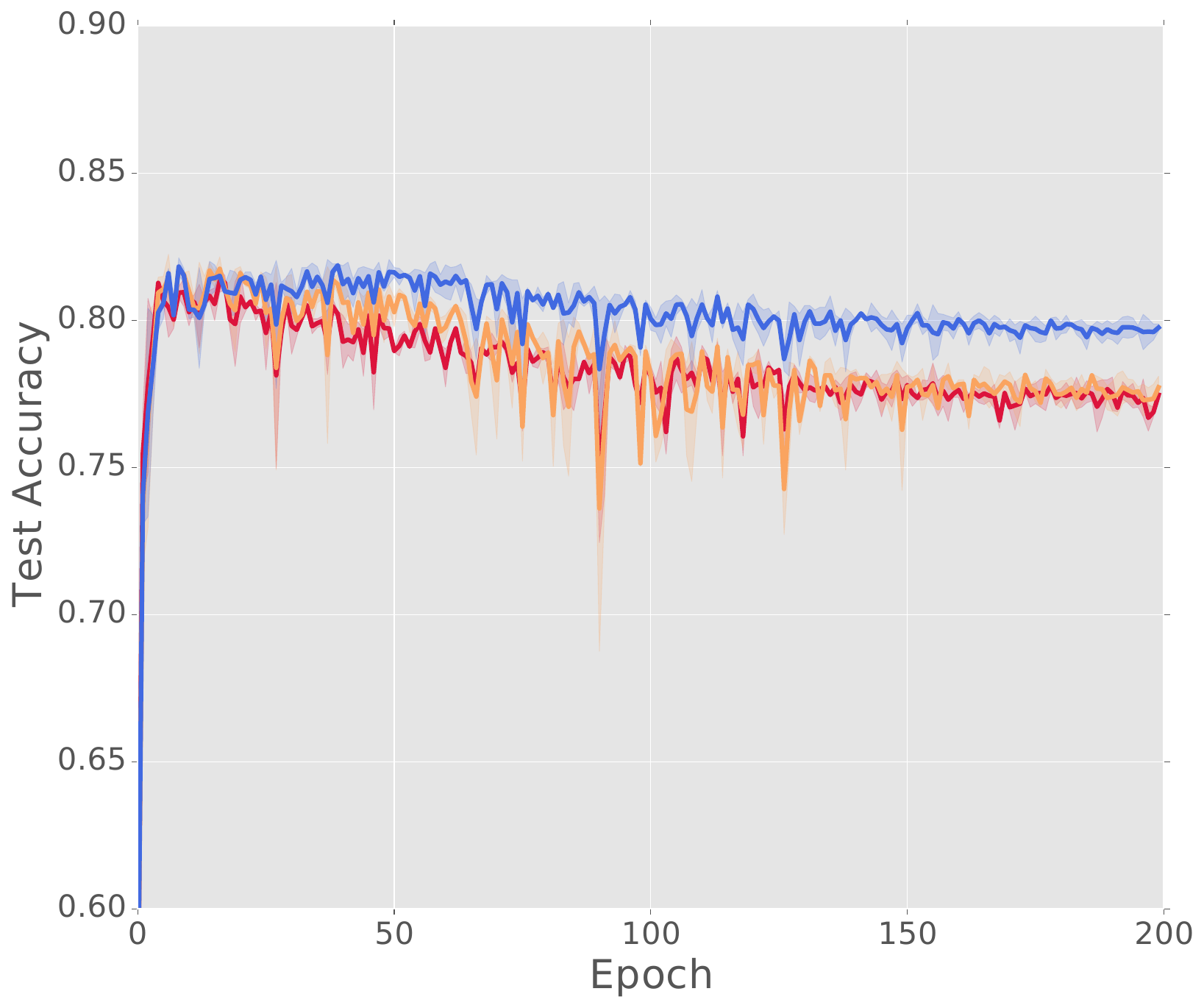}%
        \includegraphics[width=0.33\textwidth]{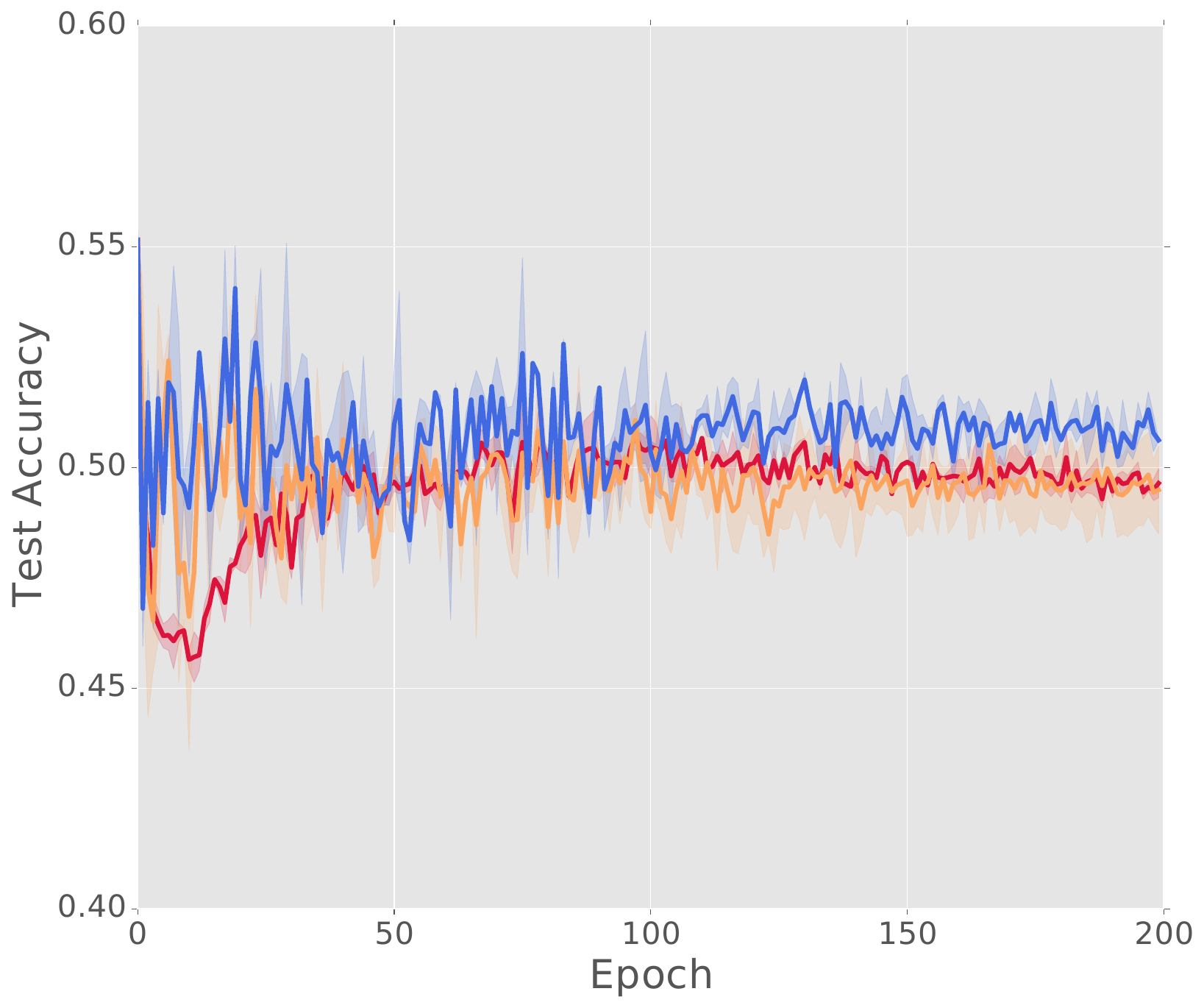}%
        \includegraphics[width=0.33\textwidth]{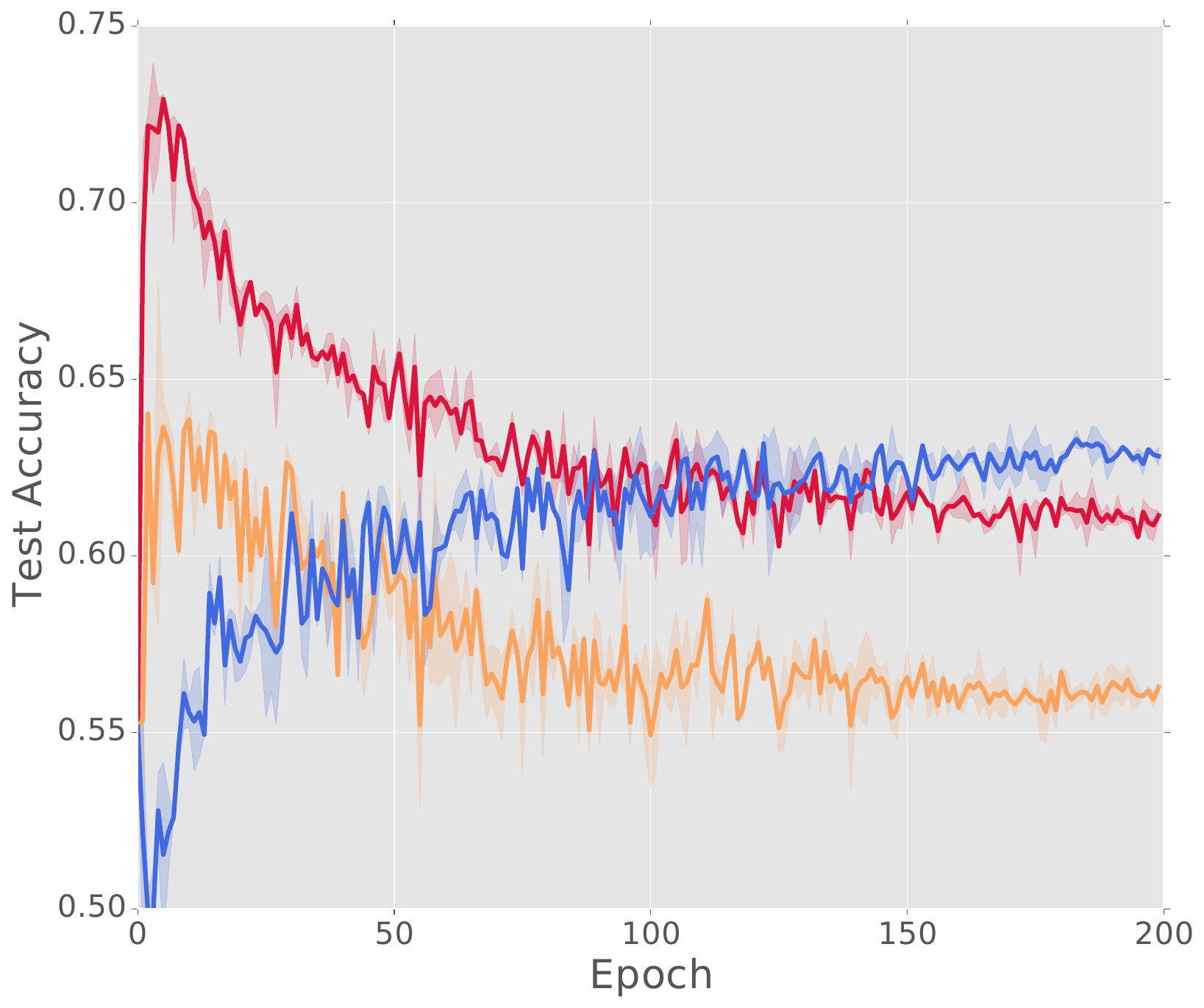}%
    \end{minipage}\\
    \vspace{-1ex}%
    \caption{Accuracy curves of training deep networks using the three learning methods in SET2.}
    \label{fig:back-corr-cifar100-news}
    \vspace{-2ex}%
\end{figure*}

Due to the limited space, the experiments on CIFAR-100 and NEWS are moved here.
The setup of CIFAR-100 is similar to CIFAR-10, but the momentum is 0.5 and lr is divided by 10 every 30 epochs for SET2 methods.
The setup of NEWS is similar to other three datasets, except $\fO$ is AdaGrad \citep{duchi2011adagrad} that automatically decays lr every mini-batch.

Figure~\ref{fig:small-loss-cifar100-news} shows the accuracy curves of the three methods in SET1 (CIFAR-100 in the top and NEWS in the bottom).
The trend in Figure~\ref{fig:small-loss-cifar100-news} is similar to the trend in Figure~\ref{fig:small-loss-mnist-cifar10} that SIGUA$\SL$ either stopped or alleviated the decrease in Standard and Self-Teach.
Especially on CIFAR-100, after a remarkable decrease in the first half, the accuracy in the second half started to increase once more, and it eventually surpassed the best accuracy that can be obtained by early stopping.
If we plot the test error rather than the test accuracy, this phenomenon is exactly an epoch-wise double descent \citep{nakkiran2020ddd}.
Figure~\ref{fig:back-corr-cifar100-news} shows the accuracy curves of the three methods in SET2 where SIGUA$\BC$ still stopped or alleviated the decrease in BC and/or nnBC.
The reason why BC suffered more under pair-45\% may be explained by the maximum and minimum elements of $T^{-1}$: when $k=10$, they are 2.101 and -1.719 under pair-45\% but 2.125 and -0.125 under symmetry-50\%.
It is interesting on CIFAR-10, nnBC and SIGUA$\BC$ under pair-45\% outperformed themselves under symmetry-20\%, which provides an evidence that the issue of negative losses can be fixed at least empirically.

\end{document}